\documentclass[11pt]{article}

\usepackage[final]{acl}

\usepackage{times}
\usepackage{latexsym}

\usepackage[T1]{fontenc}

\usepackage[utf8]{inputenc}

\usepackage{microtype}

\usepackage{inconsolata}

\usepackage{graphicx}

\usepackage{amsthm,amsmath}
\usepackage{amssymb}
\newtheorem{definition}{Definition}[section]

\usepackage{enumitem}
\setlist[itemize]{leftmargin=2pt}
\usepackage[most]{tcolorbox}
\usepackage{booktabs}

\usepackage{booktabs}
\usepackage{array}

\usepackage{caption}
\usepackage{subcaption}

\usepackage{siunitx}
\usepackage{multirow}

\usepackage{algorithm}
\usepackage{algpseudocode}

\usepackage{xcolor}

\tcbset{
  defstyle/.style={
    colback=gray!10,
    colframe=green!50!black,
    boxrule=1pt,
    arc=6pt,
    left=6pt,
    right=6pt,
    top=6pt,
    bottom=6pt
  }
}

\tcbset{
  quotestyle/.style={
    colback=gray!10,
    colframe=blue!50!black,
    boxrule=1pt,
    arc=6pt,
    left=6pt,
    right=6pt,
    top=6pt,
    bottom=6pt
  }
}

\usepackage{graphicx}

\usepackage{amsmath,amsfonts,bm}

\def\eqref#1{equation~\ref{#1}}

\def\1{\bm{1}}

\DeclareMathAlphabet{\mathsfit}{\encodingdefault}{\sfdefault}{m}{sl}
\SetMathAlphabet{\mathsfit}{bold}{\encodingdefault}{\sfdefault}{bx}{n}

\newcommand{\languageset}{\mathcal{K}}
\newcommand{\corpussize}{\Bar{D}}
\newcommand{\trainset}{\mathcal{D}}
\newcommand{\grandset}{\trainset_{\languageset}}
\newcommand{\model}{\theta}
\newcommand{\grandmodel}{\model_{\languageset}}
\newcommand{\task}{\mathcal{V}}
\newcommand{\grandtask}{\mathcal{V}_{\languageset}}
\newcommand{\law}{\mathcal{L}}
\newcommand{\ratios}{\mathbf{p}}
\newcommand{\optimalratios}{\ratios^{\star}}
\newcommand{\weights}{\mathbf{w}}
\newcommand{\familylaw}[1]{\law^{\mathbf{fam}}_{#1}}
\newcommand{\chinchillalaw}{\law^{\mathbf{chin}}}
\newcommand{\familyratio}[1]{\ratios_{#1}}
\newcommand{\mpg}{\mathcal{G}}
\newcommand{\payoff}{v^{\mpg}}
\newcommand{\sv}{\phi^{SV}}
\newcommand{\nsv}{\phi^{NSV}}
\newcommand{\improvement}{\Delta^{\mpg}_{i,j}}
\newcommand{\aggregatetransfer}{\Theta}

\newif\ifshowhighlight
\showhighlightfalse

\newcommand{\highlighted}[1]{%
  \ifshowhighlight
    \textcolor{blue}{#1}%
  \else
    #1%
  \fi
}

\title{ShapleyLaw: A Game-Theoretic Approach to Multilingual Scaling Laws}

\author{
  \textbf{Xuyang Cao\textsuperscript{1,2}\thanks{Equal contribution.}},
  \textbf{Qianying Liu\textsuperscript{1}\footnotemark[1]},
  \textbf{Chuan Xiao\textsuperscript{2}},
  \textbf{Yusuke Oda\textsuperscript{1,3}},
  \textbf{Jiayi Wang\textsuperscript{1,4}\thanks{This work was done while the author was visiting NII LLMC.},}
\\
  \textbf{Pontus Stenetorp\textsuperscript{1,4},}
  \textbf{Daisuke Kawahara\textsuperscript{1,5},}
  \textbf{Makoto Onizuka\textsuperscript{2},}
  \textbf{Sadao Kurohashi\textsuperscript{1},}
  \textbf{Shuyuan Zheng\textsuperscript{2}\thanks{Corresponding author.}}
\\
  \textsuperscript{1}NII LLMC,
  \textsuperscript{2}The University of Osaka,
  \\
  \textsuperscript{3}Nara Institute of Science and Technology,
  \textsuperscript{4}University College London,
  \textsuperscript{5}Waseda University
\\[0.4em]
 {\footnotesize
  \begin{tabular}{c}
    \texttt{\{xuyang.cao,chuanx,onizuka,zheng\}@ist.osaka-u.ac.jp \quad } \\
    \texttt{\{ying,odashi,pontus,kurohashi\}@nii.ac.jp \quad dkw@waseda.jp \quad jiaywang@cs.ucl.ac.uk}
  \end{tabular}
  }
}

\input{savespace}

\begin{document}
\maketitle

\begin{abstract}
In multilingual pretraining, the test loss of a pretrained model is heavily influenced by the proportion of each language in the pretraining data, namely the \textit{language mixture ratios}. 
Multilingual scaling laws can predict the test loss under different language mixture ratios and can therefore be used to estimate the optimal ratios. 
However, the current approaches to multilingual scaling laws do not explicitly quantify \textit{cross-lingual transfer} between individual languages, resulting in suboptimal mixture-ratio optimization.
In this paper, we consider multilingual pretraining as a cooperative game in which each language acts as a player that jointly contributes to pretraining, gaining the resulting reduction in test loss as the payoff. 
Consequently, from the perspective of cooperative game theory, we quantify the cross-lingual transfer from each language by its contribution in the game, and propose a game-theoretic multilingual scaling law called \textit{ShapleyLaw}.
Our experiments show that ShapleyLaw outperforms baseline methods in model performance prediction and language mixture optimization.\footnote{Our code and data are available at \url{https://github.com/YcTtXyLLM/ShapleyLaw}.}
\end{abstract}

\section{Introduction}

As large language models (LLMs) serve increasingly diverse users, multilingual pretraining on mixed-language corpora has become standard practice for modern foundation models~\cite{DBLP:journals/tacl/LiuGGLEGLZ20,DBLP:conf/emnlp/LinPWQFZL20,DBLP:conf/emnlp/HuangTZZSXW23,DBLP:conf/nips/0006ZCKB24,DBLP:journals/fcsc/XuHZQXYG25,DBLP:journals/tist/WangZZWMLWLXTHMHW25}.
The performance of a pretrained model on a given language $j$ depends not only on the model size $N$ and the pretraining data size $D$, but also critically on the proportion of each language $i$ in the pretraining corpus, i.e., the mixture ratios $\ratios$. 
This is because knowledge learned from language $i$ can transfer to language $j$, a phenomenon known as \textit{cross-lingual transfer}~\cite{DBLP:journals/pieee/ZhuangQDXZZXH21,DBLP:conf/nips/ConneauL19,DBLP:conf/naacl/SchusterGSL19,DBLP:conf/emnlp/PfeifferVGR20,fernandes2023scaling,DBLP:conf/acl/IsonumaT24,he-etal-2025-scaling,DBLP:conf/emnlp/WangLWRACTS25}. 
Therefore, identifying an optimal set of language mixture ratios that minimizes the test loss is a crucial problem for multilingual pretraining.

\begin{figure*}[t]
\centering
\begin{subfigure}[t]{0.48\linewidth}
    \centering
    \includegraphics[width=\linewidth]{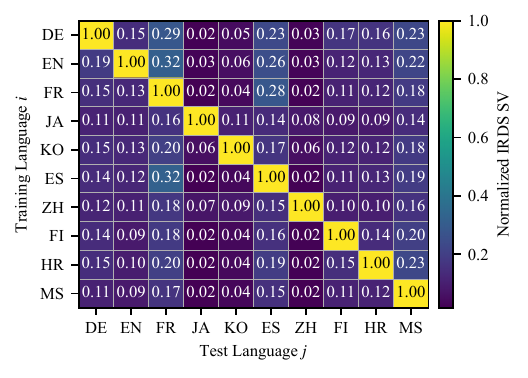}
    \caption{Cross-lingual transfer measured by the SV.}
    \label{fig:overview_a}
\end{subfigure}
\hfill
\begin{subfigure}[t]{0.48\linewidth}
    \centering
    \includegraphics[width=\linewidth]{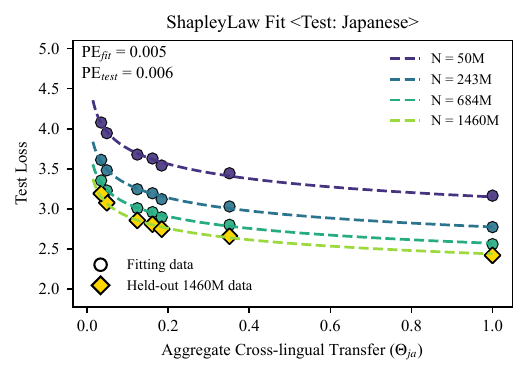}
    \caption{Test loss prediction by ShapleyLaw.}
    \label{fig:overview_b}
\end{subfigure}
\caption{
\textbf{(a)} Cross-lingual transfer effects quantified by normalized SVs estimated on a small-scale configuration ($N=50$M, $D=50$B), shown as an asymmetric matrix. In contrast, the FamilyLaw~\cite{he-etal-2025-scaling} baseline yields zero off-diagonal entries.
\textbf{(b)} Across diverse pre-training mixtures and model scales $N\in\{50\text{M},243\text{M},684\text{M},1460\text{M}\}$ with total data size $D=100$B, the proposed ShapleyLaw $\law_{j}(N,D,\aggregatetransfer_{ja})$ accurately fits and predicts Japanese performance.
}
\label{fig:overview_results}
\end{figure*}

Recently, \citet{he-etal-2025-scaling} proposed using \textit{multilingual scaling laws} to predict the test loss of pretrained models under different language mixture ratios, thereby optimizing language mixture. 
A central challenge in constructing such laws is how to measure cross-lingual transfer and relate mixture ratios to test loss. 
To simplify the complicated transfer effect among languages, the authors grouped languages by language family and assumed independence across families: \textit{cross-lingual transfer occurs only among languages within the same family.} 
Then, they constructed multilingual scaling laws for each language family, referred to as \textit{FamilyLaw}. 
However, FamilyLaw can merely be used to optimize mixture ratios across language families rather than across individual languages; the authors simply used the smoothed sampling strategy to allocate language mixture ratios within each family.

Moreover, the family-independence assumption underlying FamilyLaw does not hold. 
In our experiments, we find that cross-lingual transfer also exists between languages from different families, and that its strength varies substantially. 
For example, as Japanese and Chinese share substantial lexical overlap, although Chinese, Japanese, and Spanish belong to different language families, Japanese data improves performance on the Chinese task much more than Spanish data does (see Section~\ref{subsec:SOTA_limitation}). 
This result is consistent with the findings in~\citet{DBLP:conf/acl/LinCLLZXRHZMALN19}, which showed that the cross-lingual transfer effect from language $i$ to language $j$ largely depends on linguistic distances such as geographic distance and syntactic distance. 
By ignoring cross-family transfer, FamilyLaw may inform language mixtures that deviate substantially from the optimum.

In this work, we propose quantifying cross-lingual transfer from the perspective of cooperative game theory (CGT). 
We formulate multilingual pretraining as a cooperative game, in which each language is treated as a player and the payoff is defined by the reduction in test loss on a target evaluation language. We then use the Shapley value (SV) to measure the contribution of each training language to the target language, and interpret this contribution as cross-lingual transfer, as illustrated in Figure~\ref{fig:overview_a}.

Based on this transfer measure, we propose \textit{ShapleyLaw}, a game-theoretic multilingual scaling law that predicts test loss from model size, data size, and language mixture ratios. To capture the effect of multilingual mixture ratios on a target language, we introduce an \textit{aggregate cross-lingual transfer} quantity that summarizes the transfer effect of a mixture through SV-based transfer strengths.
In this way, ShapleyLaw extends standard scaling law formulations with a language-level characterization of cross-lingual transfer. As shown in Figure~\ref{fig:overview_b}, it accurately predicts test loss on various mixtures and model scales, enabling mixture optimization in individual languages.

We summarize our contributions as follows.
\begin{itemize}
    \item \textbf{SV-based measure of cross-lingual transfer.}
    We formulate multilingual pretraining as a cooperative game, which allows us to apply the SV to quantify the cross-lingual transfer effect across languages.
    \item \textbf{ShapleyLaw for multilingual scaling laws.}
    ShapleyLaw is the first framework to characterize the functional relationship between language-level mixture ratios $\ratios$ and test loss, thereby informing optimal mixture ratios across languages.
    \item \textbf{Empirical validation.} 
    We evaluate ShapleyLaw on multilingual pretraining tasks covering 10 languages. 
    The results show that ShapleyLaw consistently outperforms baseline methods in test loss prediction, \highlighted{mixture-ratio} optimization, and downstream task performance prediction.
\end{itemize}

\section{Preliminaries}
\label{sec:problem formulation}

\subsection{Problem Statement}
We consider pretraining a multilingual model $\grandmodel$ that supports a given set of languages $\languageset=\{1,\dots,K\}$. 
For each language $i \in \languageset$, we have a corpus of size $\corpussize_i$. 
To construct a pretraining dataset $\grandset$ of size $D$ for pretraining $\grandmodel$, for each language $i$, we sample a training subset $\trainset_i$ from the corpus according to a mixture ratio $p_i$, where the subset size is $D_i = p_i \cdot D$. 
The union of all the languages' training subsets forms the pretraining dataset, i.e., $\grandset = \bigcup_{i=1}^{K} \trainset_i$.

\paragraph{Performance prediction.} Because pretraining a large language model is computationally expensive, we aim to predict the test loss of the pretrained model $\grandmodel$ on a target evaluation task $\grandtask$ by constructing a multilingual scaling law. 
Specifically, the target task $\grandtask = \{\task_{1}, \dots, \task_{K}\}$ consists of multiple language-specific tasks, each denoted by $\task_{j}$, where $\task_{j}$ evaluates the model’s performance on language $j \in \languageset$.
For each task $\task_{j}$, we estimate a language-specific scaling law $\law_{j}(N, D, \ratios)$ to predict the loss of model $\grandmodel$ on that task, where $N$ is the model size, and $\ratios = [p_1, \dots, p_K]$ is the vector of all the language mixture ratios. 
Then, we can predict the model’s performance on the target task $\grandtask$ by aggregating all the language-specific scaling laws into a multilingual scaling law:
\begin{align*}
    \mathcal{L}(N,D,\ratios) = \sum_{j=1}^{K} w_{j}\cdot \law_{j}(N, D, \ratios)
\end{align*}
where $w_j$ denotes a weight representing the relative importance of task $\task_j$.
For example, we can determine $w_1,\dots,w_K$ by estimating the real-world distribution of language tokens across languages.

\paragraph{Mixture optimization.} In addition to predicting test loss, which is the traditional function of scaling laws, multilingual scaling laws can further be used to inform optimal mixture ratios. 
Specifically, given a model size $N$, a total data size $D$, and importance weights $\weights = [w_1, \dots, w_K]$, we can estimate the optimal mixture ratios $\optimalratios$ by solving the following optimization problem:
\begin{align}
    \label{eq:optimal_ratio}
\optimalratios = \arg\min_{\ratios}\sum_{j=1}^{K} w_{j} \cdot \law_{j}(N,D,\ratios).
\end{align}

In summary, a multilingual scaling law should (1) accurately predict test losses for different parameter configurations $(N, D, \ratios)$, and (2) enable the model pretrained with the optimized mixture ratios $\optimalratios$ to achieve a test loss as low as possible.

\subsection{Limitations of the FamilyLaw}
\label{subsec:SOTA_limitation}
\citet{he-etal-2025-scaling} proposed the current SOTA multilingual scaling law, which we refer to as FamilyLaw. 
To simplify the modeling of cross-lingual transfer, FamilyLaw groups the languages into $M$ language families and predicts the test loss $\familylaw{m}$ for each family $m\in[M]$ as follows:
\begin{align*}
    \familylaw{m}(N,D,\ratios)=(E_m\!+\!\frac{A_m}{N^{\alpha_m}}\!+\!\frac{B_m}{D^{\beta_m}})(\familyratio{m})^{-\gamma_m}
\end{align*}
where $\familyratio{m}$ is the sum of the mixture ratios for languages in family $m$, and $A_m, B_m, \alpha_m, \beta_m$ are fitting parameters.

\begin{figure}[t]
    \includegraphics[width=\columnwidth]{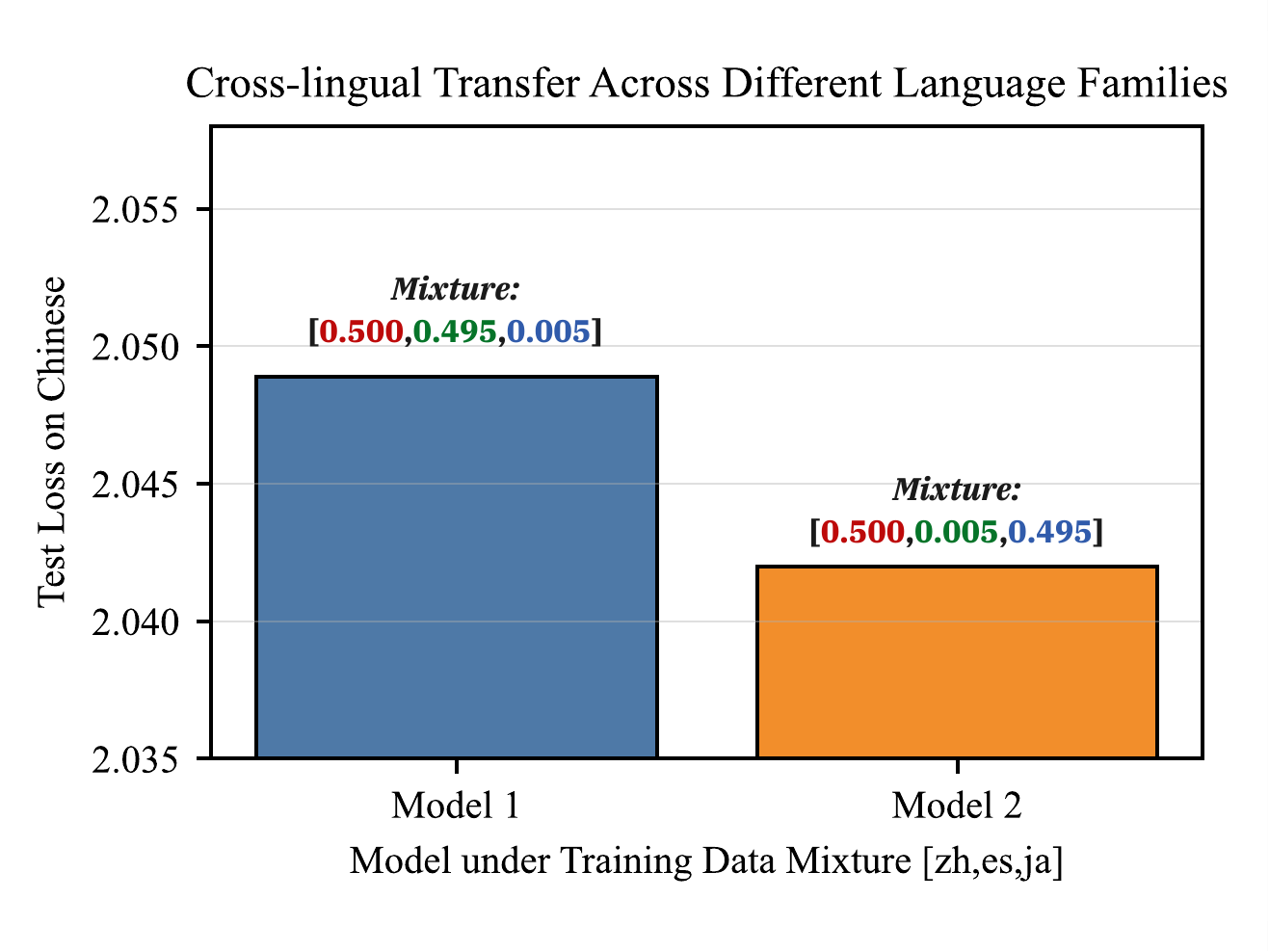}
  \caption{Consider pretraining two models for a Chinese task with training data mixtures in Chinese (zh), Japanese (\highlighted{ja}), and Spanish (es), which belong to different language families. 
  We first pretrain Model 1 by sampling Chinese, Spanish, and Japanese data with ratios $[0.500,0.495,0.005]$, respectively. 
  We then swap the mixture ratios of Japanese and Spanish to pretrain Model 2. 
  Although Models 1 and 2 share the same model size $N$=684M and data size $D$=100B, they exhibit significantly different losses on the Chinese task.}
  \label{fig:evidence}
\end{figure}

However, we find that FamilyLaw has the following three limitations.
\textbf{(1)} For performance prediction, FamilyLaw does not construct language-specific scaling laws, making it difficult to predict model performance on single-language tasks. 
\textbf{(2)} For mixture ratio optimization, FamilyLaw only optimizes mixture ratios across language families, but does not inform the optimal mixture ratios within each family (\citet{he-etal-2025-scaling} simply applied smoothed sampling to determine the intra-family ratios). 
\textbf{(3)} FamilyLaw relies on a core assumption that cross-lingual transfer occurs only within the same language family. 
However, as shown in Figure~\ref{fig:evidence}, this assumption does not hold in practice: 
Although both Japanese and Spanish belong to language families different from Chinese, Japanese data exhibits substantially stronger cross-lingual transfer for the Chinese task because Japanese and Chinese share substantial lexical overlap.

\paragraph{Our objectives.}
Considering the above limitations, we aim to develop a multilingual scaling law that (1) predicts language-specific test loss, (2) informs optimal mixture ratios across individual languages rather than language families, and (3) models cross-lingual transfer across languages.

\section{Shapley Value-Based Scaling Law}
\subsection{Multilingual Pretraining Game} 
\label{sec:problem_formulation}

The core challenge in developing multilingual scaling laws lies in quantifying cross-lingual transfers across languages. 
We address this challenge using CGT, which studies how to fairly assess the contribution of each player when multiple players jointly complete a task. 
In the context of multilingual pretraining, each language $i \in \languageset$ can likewise be viewed as a player.
These players jointly pretrain a model, which is used to complete an evaluation task $\task_j$ in a target language $j$. 
This formulation allows us to use contribution metrics from CGT to quantify the contribution of each language $i$ to the target language $j$, and thereby assess cross-lingual transfer.
We refer to this cooperative game as a \textit{multilingual pretraining game}, defined as follows.

\begin{definition}[Multilingual Pretraining Game (MPG)]
\label{def:game}
    An MPG $\mpg= (\languageset, N, D, \task)$ is specified by a language set $\languageset$, a model size $N$, a data size $D$, and a target evaluation task $\task$.
    For any language subset $S \subseteq \languageset$, the payoff $\payoff_{j}(S)$ of $S$ on any language-specific task $\task_{j} \in \task$ is the decreased test loss by pretraining $\model_{\languageset}$ using a pretraining dataset $\trainset_{S}^{uni}$, where $\trainset_{S}^{uni}$ consists of training data from languages $S$ with uniform mixture ratios.
    A contribution metric $\phi$ to the game $\mpg$ returns a $K \times K$ contribution matrix $\phi(\mpg) = \{\phi_{i,j(\mpg)}\}_{i,j\in\languageset}$ that distributes the payoff $\payoff_{j}(\languageset)$ among all the languages $\languageset$ for all task $\task_{j} \in \task$, i.e., $\forall j, \sum_{i=1}^{K}\phi_{i, j}(\mpg) = \payoff_{j}(\languageset)$.
\end{definition}

Intuitively, in the MPG $\mpg$, all the languages $\languageset$ as players participate in pretraining with the same amount of data. 
For each task $\task_j$, they gain a payoff of $\payoff_j(\languageset)$, which is the reduction in the test loss of the pretrained model. 
The contribution of language $i$ to task $\task_j$, denoted as $\phi_{i,j}(\mpg)$, measures the portion of this test loss reduction attributable to language $i$.
Consequently, we can measure the cross-lingual transfer from language $i$ to language $j$ as the contribution $\phi_{i,j}$ in MPG.

\subsection{Shapley Value for Measuring Cross-Lingual Transfer}
The Shapley value (SV) is a widely-adopted contribution metric because it is the only metric that satisfies four important properties: \textit{linearity}, \textit{efficiency}, \textit{symmetry}, and \textit{null player} (see Appendix~\ref{app:sv_properties} for details). 
In MPG, we can also employ the SV $\sv$ to compute the contribution matrix as follows:
\begin{align*}
\sv_{i,j}(\mpg) = \textstyle\sum_{S \subseteq \languageset \setminus\{i\}} W(S) \cdot \improvement(S)
\end{align*}
where $W(S) = \frac{|S|!(K-|S|-1)!}{K!}$, and $\improvement(S) = \payoff_{j}(S \cup \{i\}) - \payoff_{j}(S)$.

Intuitively, the SV measures the marginal contribution of language $i$ to the reduction in test loss on task $\task_j$, averaged over all possible subsets of languages.
Specifically, the SV (1) enumerates all language subsets $S \subseteq \languageset \setminus\{i\}$, (2) calculates the marginal improvement $\improvement(S)$ in the test loss by adding language $i$ into $S$ for pretraining, and (3) finally aggregates the improvements by language $i$ across all subsets $S$ as language $i$'s contribution to task $\task_{j}$, where $W(S)$ is the weight for aggregation.
To ensure comparability across different tasks, we further apply exponential normalization to compute the normalized SV as follows:
\begin{align*}
\nsv_{i,j}(\mpg) = \exp{\big(\sv_{i,j}(\mpg) - \max_{i'} \sv_{i',j}(\mpg) \big)}
\end{align*}
which ensures that the language $i$ with the largest contribution to task $\task_j$ has a normalized SV of $1$, while the values for other languages lie in the range $(0, 1]$. 
\highlighted{The rationale for choosing this normalization scheme is provided in Appendix~\ref{app:normalization_scheme}.}
Consequently, the normalized SV can serve as a measure of the cross-lingual transfer from any language $i$ to any language $j$.
For simplicity, we denote $\nsv_{i,j}(\mpg)$ as $\nsv_{i,j}$ in the following, when no ambiguity arises.

\subsection{Construction of ShapleyLaw}
\label{sec:shapleylaw}
Finally, we construct a multilingual scaling law, named \textit{ShapleyLaw}, based on the SV. 
Formally, given a set of parameter configurations $(N, D, \ratios)$ and a set of normalized SVs $\{\nsv_{i,j}\}$ that represent cross-lingual transfers among languages in $\languageset$, ShapleyLaw predicts test loss on each task $\task_j$ as:
\begin{align*}
    \law_{j}(N, D, \ratios \mid \{\nsv_{i,j}\}) = \chinchillalaw_{j}(N, D) \cdot (\aggregatetransfer_j)^{-\gamma_j}
\end{align*}
where $E_j, A_j, B_j, \alpha_j, \beta_j, \gamma_j$ are fitting parameters; 
$\chinchillalaw_{j}(N, D) = E_j \!+\! \frac{A_j}{N^{\alpha_j}} \!+\! \frac{B_j}{D^{\beta_j}}$ corresponds to the Chinchilla scaling law~\citep{HoffmannTraining22} that does not consider multilingual mixture ratios;
and $\aggregatetransfer_j = \sum_{i\in \languageset} p_i \cdot \nsv_{i, j}$ measures the \textit{aggregate cross-lingual transfer} from languages $\languageset$ to language $j$ weighted by their mixture ratios.
\highlighted{For a fixed model size $N$ and data budget $D$, different multilingual mixtures can yield different losses on task $\task_j$ due to differences in their cross-lingual transfer effects. Accordingly, $\chinchillalaw_j(N,D)$ captures the base scaling behavior, while $\aggregatetransfer_j$ characterizes the effective language-specific support induced by the data mixture; their multiplicative interaction models cross-lingual transfer as modulating how effectively a fixed scale budget benefits the target language. For example, in Figure~\ref{fig:evidence}, the aggregate transfer to Chinese is computed as $\aggregatetransfer_{\mathrm{zh}}=\sum_i p_i\nsv_{i,\mathrm{zh}}$. Swapping the mixture ratios of Japanese and Spanish changes $\aggregatetransfer_{\mathrm{zh}}$ according to their respective transfer contributions to Chinese, thereby yielding different Chinese losses despite identical $N$ and $D$.}

\begin{figure*}[t]
  \centering
  \includegraphics[width=0.48\linewidth]{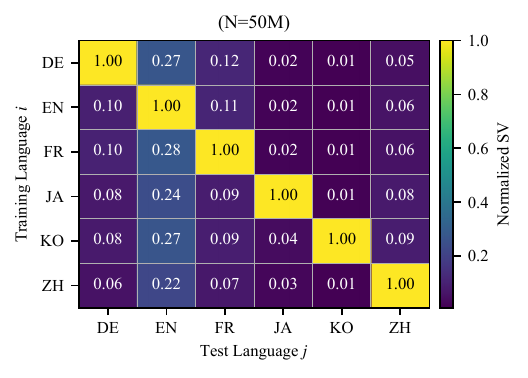}
  \hfill
  \includegraphics[width=0.48\linewidth]{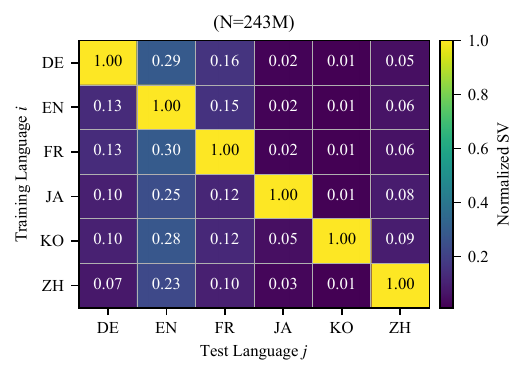}
    
  \caption{\textbf{Comparison of Shapley values across model sizes.}
  \textbf{Left}: normalized Shapley values estimated with $N=50$M and $D=50$B.
  \textbf{Right}: normalized Shapley values estimated with $N=243$M and $D=50$B.
  The two heatmaps exhibit highly consistent cross-lingual transfer patterns across model sizes.}
  \label{fig:sv_transfer_heatmap_N}
\end{figure*}



\paragraph{SV transferability.}
\highlighted{A key assumption of ShapleyLaw is that the SVs used to quantify cross-lingual transfer are transferable across model sizes.
Under this assumption, the SVs estimated from a relatively small model can be reused as fixed transfer coefficients when fitting scaling laws for larger models.
To empirically verify this property, we conduct a controlled comparison across two model sizes, $N=50$M and $N=243$M, while keeping the pretraining data size fixed at $D=50$B.
We consider six language sets, $\mathcal{K}=\{\mathrm{de}, \mathrm{en}, \mathrm{fr}, \mathrm{ja}, \mathrm{ko}, \mathrm{zh}\}$.
For each model size, we construct a multilingual pretraining game over the same language set and estimate the normalized SV matrix
$\nsv \in \mathbb{R}^{|\mathcal{K}|\times|\mathcal{K}|}$,
where each entry $\nsv_{i,j}$ measures the transfer contribution from training language $i$ to evaluation language $j$.
We then compare the two normalized SV matrices obtained at different model sizes. We use cosine similarity and Pearson correlation coefficient as the comparison metrics.
Specifically, we vectorize the two SV matrices in the same language order and compute their similarity over all language pairs.}


\highlighted{Figure~\ref{fig:sv_transfer_heatmap_N} shows the normalized SV matrices estimated from the $50$M and $243$M models, which exhibit highly consistent cross-lingual transfer structures. In both cases, self-transfer dominates, while the off-diagonal entries reveal similar asymmetric patterns, such as strong transfer from English to several European languages and consistent interactions among Japanese, Korean, and Chinese. Quantitatively, the two vectorized SV matrices achieve a cosine similarity and Pearson correlation of $0.999$ and $0.999$, respectively; after excluding the diagonal entries, they remain high at $0.995$ and $0.989$. These results demonstrate that the cross-lingual transfer structure captured by SVs remains highly stable as the model scales from $50$M to $243$M parameters.}
\highlighted{Overall, these results support the transferability of SVs across model sizes, substantially reducing the computational cost of estimating transfer effects while preserving the transfer structure required for accurate multilingual scaling-law fitting.}

\highlighted{Because SV-based cross-lingual transfer exhibits transferability,} we can construct an MPG $\mpg = (\languageset, N', D', \task)$ using a small model size $N'$ and data size $D'$ to compute the normalized SVs. These normalized SVs can then be treated as fixed parameters when constructing ShapleyLaw for larger model sizes $N \gg N'$ and data sizes $D \gg D'$.
As shown in Figure~\ref{fig:overview_a}, we compute the normalized SVs under the setting $N' = 50\text{M}$ and $ D' = 50\text{B}$;
ShapleyLaw accurately fits the test losses across settings with $N \in \{243\text{M}, 684\text{M}, 1460\text{M}\}$ and $D=100\text{B}$.

In addition to predicting language-specific test loss, ShapleyLaw can also be used to estimate \emph{optimal mixture ratios} by solving the optimization problem in Formula~\ref{eq:optimal_ratio} (see Appendix~\ref{app:optimal_mixture}). 
Therefore, ShapleyLaw achieves our three objectives.

\paragraph{SV approximation.}
Since the SV $\sv_{i,j}$ enumerates all non-empty language subsets $S\subseteq \languageset$ to evaluate their payoff $\payoff_j(S)$, we need to pretrain $2^{K} - 1$ models for exact SV computation, which is NP-hard. 
To accelerate SV computation, we employ the In-Run Data Shapley (IRDS) approximation technique~\cite{DBLP:conf/iclr/WangMS025} (see Appendix~\ref{app:IRDSsv}) in our experiments, which pretrains only one model to approximate the SVs. 
With IRDS, ShapleyLaw achieves $O(1)$ computational complexity and is therefore a tractable multilingual scaling law that can scale up to dozens or even hundreds of languages.
Our experiments in Appendix~\ref{app:approx_sv_performance} show that IRDS accurately approximates the exact SV.

\begin{figure*}[t]
  \centering
  \includegraphics[width=0.32\linewidth]{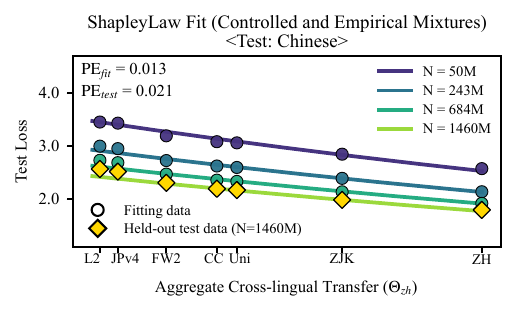}
  \hfill
  \includegraphics[width=0.32\linewidth]{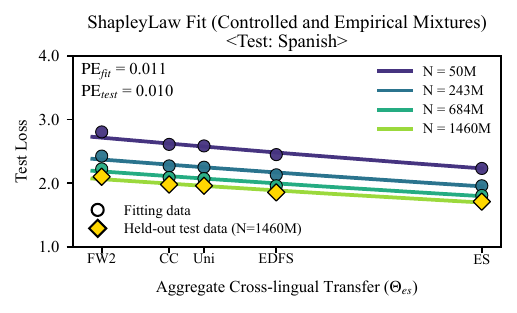}
  \hfill
  \includegraphics[width=0.32\linewidth]{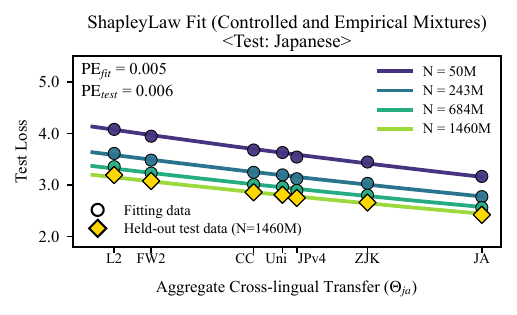}
  
  \vspace{0.5em}
  
  \includegraphics[width=0.32\linewidth]{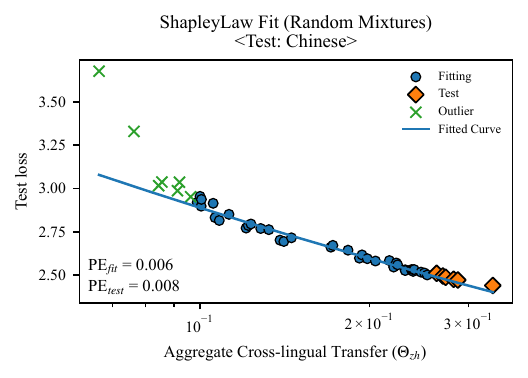}
  \hfill
  \includegraphics[width=0.32\linewidth]{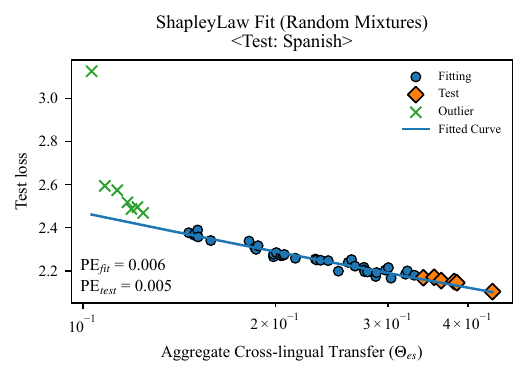}
  \hfill
  \includegraphics[width=0.32\linewidth]{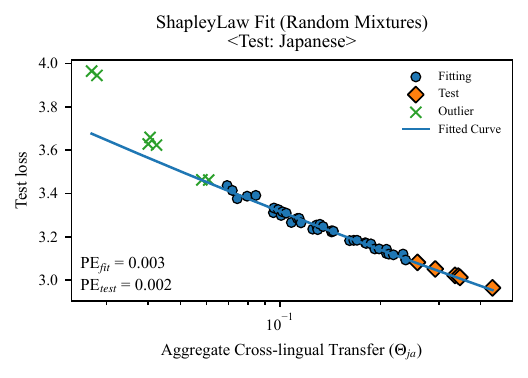}

  \vspace{0.5em}
  
  \includegraphics[width=0.32\linewidth]{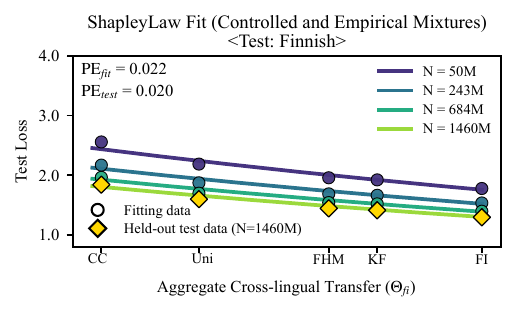}
  \hfill
  \includegraphics[width=0.32\linewidth]{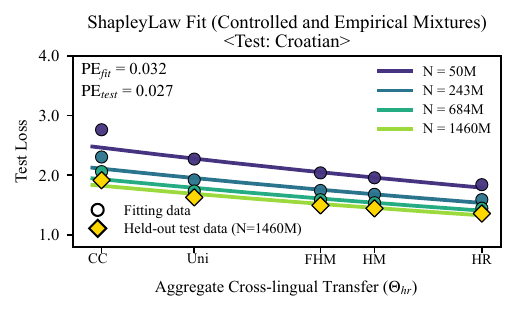}
  \hfill
  \includegraphics[width=0.32\linewidth]{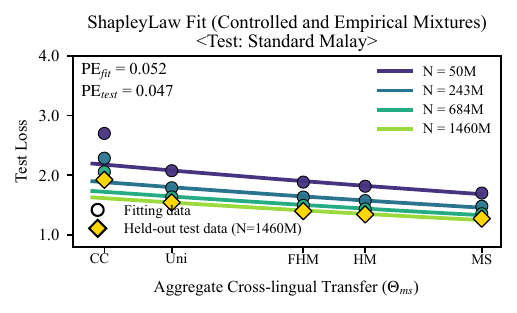}
  \caption{\textbf{Fitting ShapleyLaw across multilingual mixtures and target languages.} \textbf{Top row}: fits with respect to aggregate cross-lingual transfer $\Theta$ on curated multilingual mixtures for Chinese (ZH), Spanish (ES), and Japanese (JA) across four model scales $\{50\text{M}, 243\text{M}, 684\text{M}, 1460\text{M}\}$; circles denote fitting points and yellow diamonds denote held-out evaluation points. \textbf{Middle row}: fits under randomly sampled mixtures at model scale $N=243$M; circles indicate fitting points, yellow diamonds indicate held-out points, and orange crosses denote removed outliers. \textbf{Bottom row}: fits with respect to aggregate cross-lingual transfer $\Theta$ on curated multilingual mixtures for low-resource languages Finnish (FI), Croatian (HR), and Standard Malay (MS).}

  \label{fig:extrapolation}
\end{figure*}

\section{Experiments}

\paragraph{Setup.}
We describe the basic experimental setup here. 
More details are provided in Appendix~\ref{app:setup}.
We study $K=10$ languages from eight language families, as summarized in Table~\ref{table:corpus}.
We sample $0.1\%$ of the corpora as a fixed test set for evaluating all pretrained models. 
We consider four model scales with non-embedding parameter counts $N \in \{50\mathrm{M}, 243\mathrm{M}, 684\mathrm{M}, 1460\mathrm{M}\}$, following the open-source LLM-jp configurations. All models share the same architecture family and tokenizers.
We consider the cross-entropy (CE) loss for evaluation.
The SVs are computed in an MPG with $N=50\text{M}$ and $D=50\text{B}$ across all experiments.


\paragraph{Research questions.}
Our experiments address the following research questions.
\textbf{Q1} (Performance Prediction): Can ShapleyLaw more accurately predict the model’s test loss?
\textbf{Q2} (Mixture Optimization): Can ShapleyLaw achieve \highlighted{better language mixtures}?
\textbf{Q3} (Downstream Tasks): Can ShapleyLaw’s ability to predict model performance and optimal mixture ratios generalize to downstream tasks?
\textbf{Q4} (SV Approximation): How is the fitting performance of ShapleyLaw based on approximate SVs? (See Appendix~\ref{app:approx_sv_performance}.)


\begin{figure}[t]
  \includegraphics[width=\linewidth]{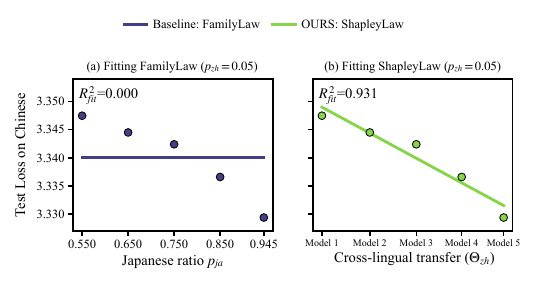}
  \caption{\textbf{Controlled comparison between FamilyLaw and ShapleyLaw on a Chinese evaluation task.} We pretrain multilingual models with $N=50$M and $D=100$B using Chinese, Japanese, and Spanish corpora. Starting from the mixture $[0.05, 0.55, 0.40]$, we vary the Spanish--Japanese proportions up to $[0.05, 0.945, 0.05]$ while keeping the Chinese ratio fixed. Performance is evaluated on the Chinese test set. \textbf{Left:} FamilyLaw, where x-axis stands for language ratio, yields a poor fit. \textbf{Right:} ShapleyLaw, where the x-axis represents different Chinese aggregate transfer $\Theta$, fits accurately.}

  \label{fig:baseline_compare}
\end{figure}

\subsection{Experiments on Performance Prediction}
\label{sec:shapley_fit}

\paragraph{Experiment design.}
We pretrain models with $N \in \{50\mathrm{M}, 243\mathrm{M}, 684\mathrm{M}, 1460\mathrm{M}\}$ and $D=100$B. 
At each model scale, we consider both \textit{controlled mixtures} and \textit{empirical mixtures}. 
Controlled mixtures isolate the effect of language mixtures under fixed ratios, including monolingual training, uniform mixtures over selected languages (e.g., zh--ja--ko, fi-hr-ms and en--de--fr--es, abbreviated as \textit{ZJK}, \textit{FHM} and \textit{EDFS}), and the uniform mixture over all languages (\textit{Uni}).
Empirical mixtures mirror multilingual data distributions adopted in prior LLM studies, including the language distributions of Common Crawl\footnote{https://commoncrawl.org} and FineWeb-2~\cite{penedo2025fineweb2pipelinescale}, the English-centric multilingual model LLaMA-2~\cite{DBLP:journals/corr/abs-2307-09288}, and the Japanese-focused pretraining corpus LLM-JP-V4\footnote{https://gitlab.llm-jp.nii.ac.jp/datasets/llm-jp-corpus-v4}~\cite{DBLP:journals/corr/abs-2407-03963} (abbreviated as \textit{CC}, \textit{FW2}, \textit{L2}, and \textit{JPv4}). To evaluate the fitting performance, we also use the normalized MAE to quantify the prediction error (PE).

In addition, we examine 50 randomly sampled language mixtures under $N=243\mathrm{M}$ to further evaluate generalization to unseen mixture ratios. 
To ensure reliable fitting, we exclude those random mixtures where a training language $i$ has subset size $D_i \leq 1.5$B. 
This regime corresponds to extremely low data availability for language $i$, causing optimization to become unstable even for $N=243$M models. 
This threshold remains practically relevant: in FineWeb-scale corpora, a language budget of roughly $1.5\text{B}$ tokens already extends to many low-resource languages considered in current multilingual LLM studies, including Basque and Armenian.
We then split the remaining runs into fitting and held-out evaluation subsets.

\paragraph{Results.}
Figure~\ref{fig:extrapolation} shows the fitted results for Chinese, Spanish, and Japanese. On curated multilingual mixtures across four model scales, ShapleyLaw achieves strong held-out performance, with $\mathrm{PE}_{test}=0.021$ for Chinese, $0.010$ for Spanish, and $0.006$ for Japanese ($R^2=0.990$, $0.997$, $0.996$). On randomly sampled mixtures at $N=243$M, the fitted curves remain smooth with respect to $\Theta$, with training-set fits of $\mathrm{PE}_{test}=0.008$, $0.005$, and $0.002$ for Chinese, Spanish, and Japanese ($R^2=0.975$, $0.942$, $0.987$), respectively. These results indicate that the proposed method makes stable predictions across sampling regimes and typologically diverse languages. Additional results for French, German and Korean are provided in Appendix~\ref{app:performance_predict}. 

We next compare ShapleyLaw with the family-level baseline FamilyLaw in a controlled setting that isolates cross-family transfer. As shown in Figure~\ref{fig:baseline_compare}, we run experiments on the \textit{ZJS} mixture. We fix the Chinese mixture ratio at $p_{\mathrm{zh}}=0.05$ and vary only the Japanese--Spanish proportions, keeping the total token budget fixed. Under FamilyLaw, both Japanese and Spanish lie outside the Chinese language family, so changing their ratios should not affect Chinese performance; accordingly, the fitted relation is nearly constant and fails to explain the observed variation ($R^2 \leq 0$). In contrast, ShapleyLaw captures the cross-language transfer and fits the loss on Chinese much more accurately ($R^2=0.931$). This result shows that family-level aggregation is too coarse to model important cross-lingual interactions in multilingual pretraining, such as shared script and lexical overlap.

\highlighted{Moreover, Appendix~\ref{app:leave_one_language_out} further demonstrates that ShapleyLaw generalizes well to a held-out language when a related language from the same family is retained in the fitting set.
}

\begin{table*}[t]
\centering
\small
\begin{subtable}[t]{0.49\textwidth}
\vspace{0pt}
\centering
\setlength{\tabcolsep}{3.2pt}
\renewcommand{\arraystretch}{1.1}

\resizebox{\linewidth}{!}{
\begin{tabular}{
l
*{10}{S[table-format=2.1]}
@{\hspace{6pt}}
S[table-format=2.3]
}
\toprule
\multirow{2}{*}[-0.4ex]{\normalsize{Strategy}}
& \multicolumn{10}{c}{\textbf{Language Mixture Ratio (\%)}} 
& {\textbf{CE Loss $\downarrow$}} \\
\cmidrule(lr){2-11}
& {en} & {de} & {fr} & {es} & {zh} & {ja} & {ko} & {fi} & {hr} & {ms}
& {$\mathcal{L}$} \\
\midrule
Uniform
& 10.0 & 10.0 & 10.0 & 10.0 & 10.0 & 10.0 & 10.0 & 10.0 & 10.0 & 10.0
& 24.440 \\

{$\alpha = 0.5$}
& 13.2 & 14.5 & 12.6 & 13.6 & 19.1 & 11.4 & 4.9 & 4.7 & 3.7 & 2.4
& 24.683 \\

Common Crawl
& 61.4 & 8.3 & 6.4 & 6.4 & 7.2 & 8.2 & 1.2 & 0.5 & 0.3 & 0.1
& 25.491 \\

FineWeb-2
& 80.6 & 3.3 & 2.3 & 2.7 & 6.7 & 3.1 & 0.5 & 0.4 & 0.3 & 0.1
& 27.133 \\

FamilyLaw
& 14.1 & 14.1 & 1.8 & 1.8 & 12.1 & 11.6 & 10.2 & 12.6 & 11.0 & 10.7
& 24.749 \\

ODM
& 1.5 & 30.0 & 5.1 & 7.2 & 6.0 & 30.0 & 1.5 & 1.5 & 15.7 & 1.5
& 25.544 \\

DML
& 1.5 & 36.7 & 4.7 & 17.5 & 5.9 & 5.0 & 1.5 & 5.3 & 8.9 & 13.0
& 25.330 \\

ATLAS
& 1.5 & 1.5 & 5.1 & 1.7 & 3.4 & 28.0 & 23.3 & 12.4 & 12.5 & 10.5
& 25.353 \\

\midrule
\textbf{ShapleyLaw}
& 26.0 & 13.5 & 1.5 & 1.5 & 20.2 & 13.4 & 10.3 & 5.3 & 6.8 & 1.5
& \textbf{24.108} \\

\bottomrule
\end{tabular}
}

\label{tab:mixture_strategies_unweighted}
\end{subtable}
\hfill
\begin{subtable}[t]{0.49\textwidth}
\vspace{0pt}
\centering
\setlength{\tabcolsep}{3.2pt}
\renewcommand{\arraystretch}{1.1}

\resizebox{\linewidth}{!}{
\begin{tabular}{
l
*{10}{S[table-format=2.1]}
@{\hspace{6pt}}
S[table-format=1.3]
}
\toprule
\multirow{2}{*}[-0.4ex]{\normalsize{Strategy}}
& \multicolumn{10}{c}{\textbf{Language Mixture Ratio (\%)}} & {\textbf{CE Loss $\downarrow$}} \\
\cmidrule(lr){2-11}
& {en} & {de} & {fr} & {es} & {zh} & {ja} & {ko} & {fi} & {hr} & {ms}
& {$\mathcal{L}$} \\
\midrule
Uniform
& 10.0 & 10.0 & 10.0 & 10.0 & 10.0 & 10.0 & 10.0 & 10.0 & 10.0 & 10.0
& 11.985 \\

{$\alpha = 0.5$}
& 13.2 & 14.5 & 12.6 & 13.6 & 19.1 & 11.4 & 4.9 & 4.7 & 3.7 & 2.4
& 12.180 \\

Common Crawl
& 61.4 & 8.3 & 6.4 & 6.4 & 7.2 & 8.2 & 1.2 & 0.5 & 0.3 & 0.1
& 12.715 \\

FineWeb-2
& 80.6 & 3.3 & 2.3 & 2.7 & 6.7 & 3.1 & 0.5 & 0.4 & 0.3 & 0.1
& 13.613 \\

FamilyLaw
& 9.9 & 9.9 & 1.5 & 1.5 & 11.6 & 8.6 & 9.3 & 17.4 & 14.9 & 15.4
& 12.153 \\

ODM
& 1.5 & 30.0 & 3.7 & 9.1 & 5.7 & 30.0 & 1.5 & 1.5 & 15.5 & 1.5
& 12.592 \\

DML
& 1.5 & 37.9 & 4.5 & 17.8 & 8.7 & 17.8 & 2.3 & 2.0 & 3.5 & 3.9
& 12.304 \\

ATLAS
& 1.5 & 1.5 & 2.1 & 1.9 & 3.3 & 29.2 & 19.0 & 14.6 & 13.9 & 13.0 
& 12.518 \\

\midrule
\textbf{ShapleyLaw}
& 18.1 & 15.4 & 1.5 & 1.5 & 19.7 & 9.3 & 9.6 & 10.3 & 12.0 & 2.6
& \textbf{11.813} \\

\bottomrule
\end{tabular}
}
\label{tab:mixture_strategies_normalized}
\end{subtable}

\caption{Comparison of mixture optimization strategies and their CE loss under the Unweighted Sum (\textbf{Left}) and Normalized Sum (\textbf{Right}) settings. }
\label{tab:mixture_strategies}

\vspace{-2mm}
\end{table*}

\subsection{Experiments on Mixture Optimization}
\label{subsec:optimal_mixture}

\begin{figure*}[t]
    \includegraphics[width=0.32\linewidth]{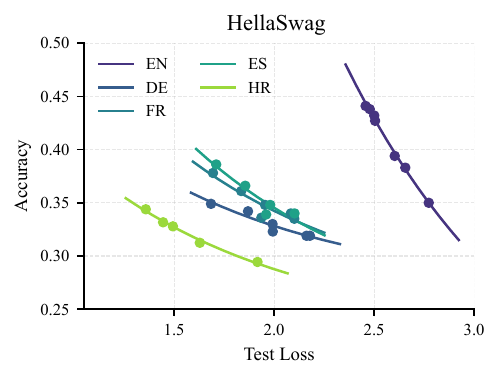}\hfill
    \includegraphics[width=0.32\linewidth]{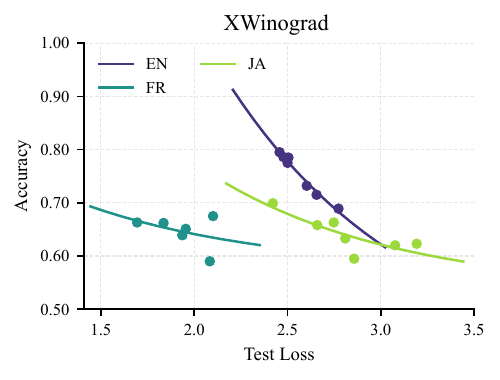}
    \includegraphics[width=0.32\linewidth]{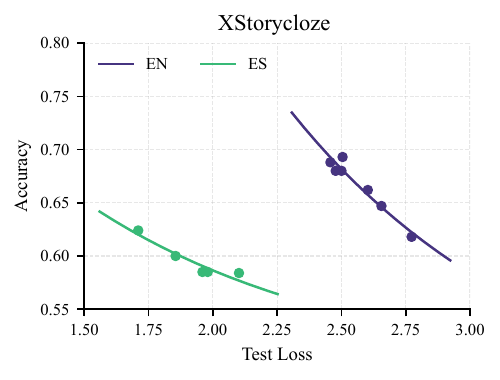}
  \caption {\textbf{Fitting Relationship between Downstream Performance and Test CE Loss.}
    Experiments are conducted on six downstream tasks for English, German, French, Spanish, Croatian, and Japanese. Across all tasks, we observe strong \emph{negative correlations} ($|r| > 0.90$, with an average $R^{2}=0.933$), indicating that lower test loss consistently corresponds to higher downstream performance.} 
  \label{fig:ce_vs_score}
\end{figure*}

\paragraph{Experiment design.}
We compare the ShapleyLaw-based mixture optimization against \highlighted{eight} representative baselines chosen to reflect the main strategies used in multilingual data allocation, including \textit{Uniform Sampling}, \textit{Smoothed Sampling}, \textit{FamilyLaw Optimization}, \highlighted{\textit{ODM}~\cite{NEURIPS2025_bc1d640f}, \textit{DML}~\cite{DBLP:conf/iclr/YeLSZZQ25}, \textit{ATLAS}~\cite{ICLR2026_35c26a81},} and \textit{Empirical Mixtures} (see Appendix~\ref{app:setup} for baseline introduction and Appendix~\ref{app:clarification_sampling_baselines} for baseline selection clarification).  
We pretrain a model ($N=243$M and $D=100$B) using the mixture ratios obtained by each method.

We consider two settings of $w_j$ for weighting the importance of each language-specific task $\task_j$. 
\textbf{Unweighted Sum:} Each task $\task_j$ receives equal weight, so the objective is to minimize the average multilingual loss. 
\textbf{Normalized Sum:} Each task $\task_j$ is weighted by the inverse of its monolingual loss. 
This ensures that intrinsically harder languages do not dominate the aggregate objective in Formula~\ref{eq:optimal_ratio} and provides a more balanced evaluation of multilingual trade-offs across languages.

\paragraph{Results.} 
Table~\ref{tab:mixture_strategies} shows the optimized mixtures and their resulting CE loss. Under both preference settings, the mixtures obtained from ShapleyLaw outperform all baselines in the final weighted loss. The gains are especially meaningful because they are obtained under actual pretraining rather than only from the fitted objective, indicating that the learned law transfers effectively from prediction to optimization. By contrast, the mixture obtained from FamilyLaw is consistently less effective in the per-language setting, suggesting that family-level aggregation is too coarse to capture the trade-offs needed for fine-grained multilingual allocation.


\subsection{Experiments on Downstream Tasks}
\label{sec:results_downstream}

\paragraph{Experiment design.}
We focus on three multilingual benchmarks that cover complementary NLU abilities: \textbf{HellaSwag}~\cite{zellers-etal-2019-hellaswag,lai-etal-2023-okapi} for commonsense reasoning through contextually appropriate ending prediction, \textbf{XStoryCloze}~\cite{DBLP:conf/emnlp/LinMAWCSOGBDPSK22,mostafazadeh-etal-2017-lsdsem} for cross-lingual story ending prediction, and \textbf{XWinograd} for cross-lingual coreference resolution~\cite{DBLP:conf/acl/TikhonovR21}.

We evaluate pretrained models under the same setting as Section~\ref{sec:shapley_fit}, using $N=1.46$B and $D=100$B, and the same multilingual mixture families (controlled and empirical mixtures). We report 5-shot accuracy on the standard benchmark splits. Specifically, we evaluate HellaSwag in \{en, fr, es, de, hr\}, XWinograd in \{en, fr, ja\}, and XStoryCloze in \{en, es\}. For each task-language pair, we compare downstream accuracy against the corresponding pretraining test cross-entropy on the same language, measured on the held-out test corpus. This yields one calibration line per task-language pair.

\paragraph{Results.}
We fit a simple calibration model $s = f(\ell) = a\cdot  g(\ell) + b$
where $\ell$ is the test loss, $s$ is downstream accuracy, and $a,b$ are fitting parameters~\cite{DBLP:journals/tmlr/ChenH0W0025}. 
We choose $g(\ell) = \exp(-\ell)$ as the transformation function $g(\cdot)$ (see more details in Appendix~\ref{app:downstream_task}). Figure~\ref{fig:ce_vs_score} shows the resulting fits for the three multilingual benchmarks, with one line per task-language pair and points colored by language. Across all evaluated languages, lower pretraining loss consistently corresponds to higher downstream accuracy, and this relationship is generally well captured by a transformed linear fit. The fitted results show that this trend holds not only within a single benchmark, but across multilingual commonsense reasoning, story completion, and coreference resolution tasks. Table~\ref{tab:downstream_fit_r2} reports the corresponding $R^2$ values for each task-language pair.

\begin{table}[t]
\centering
\footnotesize
\setlength{\tabcolsep}{3.2pt}
\renewcommand{\arraystretch}{1.08}
\begin{tabular}{lcccccc}
\toprule
Task & de & en & fr & es & ja & hr \\
\midrule
HellaSwag   & 0.918 & 0.996 & 0.864 & 0.897 & --    & 0.992 \\
XWinograd   & --    & 0.980 & 0.975 & --    & 0.675 & --    \\
XStorycloze & --    & 0.947 & --    & 0.905 & --    & --    \\
\bottomrule
\end{tabular}
\caption{Corresponding $R^{2}$ values for the fitted relationship between downstream performance and test CE loss for each task–language pair. English achieves the highest $R^{2}$, while Japanese is the lowest. The results consistently show a negative correlation between downstream performance and test CE loss.}
\label{tab:downstream_fit_r2}
\vspace{-2mm}
\end{table}

These results show that for the multilingual NLU benchmarks considered here, language-specific pretraining loss remains strongly informative for downstream accuracy across diverse mixtures. As ShapleyLaw predicts this loss accurately across language mixtures, it provides a useful intermediate signal for downstream-oriented mixture design.

\paragraph{Optimal Ratio Downstream Experiment for HellaSwag}
We further evaluated downstream mixture optimization for HellaSwag in \{en, fr, es, de, hr\}. To assess the effectiveness of ShapleyLaw on downstream tasks, we pretrained six models using different sampling strategies and evaluated them on HellaSwag. The overall performance was measured by the average accuracy across the five languages.

\begin{table}[t]
    \centering
    \small

    \setlength{\tabcolsep}{3.2pt}
    \renewcommand{\arraystretch}{1.1}
    
    \resizebox{\linewidth}{!}{
    \begin{tabular}{
    l
    *{5}{S[table-format=2.1]}
    @{\hspace{6pt}}
    S[table-format=2.3]
    }
    \toprule
    \multirow{2}{*}[-0.4ex]{\normalsize{Strategy}}
    & \multicolumn{5}{c}{\textbf{Language Mixture Ratio (\%)}} 
    & {\textbf{Overall Acc. $\uparrow$}} \\
    \cmidrule(lr){2-6}
    & {en} & {de} & {fr} & {es} & {hr} 
    &  {HellaSwag} \\
    \midrule
    Uniform
    & 20.0 & 20.0 & 20.0 & 20.0 & 20.0 
    & 0.343 \\

    {$\alpha=0.5$}
    & 25.2 & 22.9 & 21.9 & 23.6 & 6.4 
    & 0.344 \\
    
    Common Crawl
    & 73.9 & 10.0 & 7.7 & 7.8 & 0.6 
    & 0.354 \\
    
    FineWeb-2
    & 90.3 & 3.7 & 2.5 & 3.0 & 0.5 
    & 0.355 \\
    
    FamilyLaw
    & 31.7 & 31.7 & 4.1 & 4.1 & 28.4 
    & 0.346 \\
    
    ODM
    & 0.1 & 88.2 & 0.1 & 11.5 & 0.1 
    & 0.321 \\
    
    DML
    & 0.1 & 21.1 & 0.1 & 78.6 & 0.1 
    & 0.331 \\
    
    ATLAS
    & 0.1 & 19.5 & 38.4 & 15.7 & 26.3 
    & 0.338 \\
    
    \midrule
    \textbf{ShapleyLaw}
    & 95.7 & 0.1 & 0.1 & 0.1 & 4.0 
    & \textbf{0.356} \\
    
    \bottomrule
    \end{tabular}
    }
    
    \hfill
    \caption{Comparison of mixture optimization strategies and average HellaSwag downstream-task accuracies across five languages under the unweighted-sum setting. ShapleyLaw achieves the best performance, with an average accuracy of 0.356.}
    \label{tab:downstream_mixture_strategies}
\end{table}

\paragraph{Results.}
As shown in Table~\ref{tab:downstream_mixture_strategies}, ShapleyLaw achieves the highest overall accuracy on the multilingual HellaSwag task, reaching 0.356 under the unweighted-sum setting. This result outperforms all baseline strategies. Notably, ShapleyLaw assigns most of the mixture to English while retaining a small proportion of Croatian, suggesting that the optimal downstream mixture ratio is highly English-dependent. These results demonstrate that ShapleyLaw can identify more effective language mixture ratios for downstream performance than baseline strategies, including FamilyLaw and ATLAS.






\section{Related Work}

Neural scaling laws describe predictable relationships between model performance and scale variables such as model parameters, training tokens, and compute~\cite{DBLP:journals/corr/abs-2001-08361,HoffmannTraining22}, but standard formulations do not account for \emph{data composition}. Recent work extends scaling-law analysis to data mixtures in domain-specific pretraining~\cite{DBLP:conf/nips/QueLZZQMDBWZ00W24,NEURIPS2025_bc1d640f}. These works highlight the importance of data composition, but they do not provide a multilingual scaling law that directly optimizes \emph{language-level} mixture ratios. In multilingual settings, prior work either models mixture ratios at the family level~\cite{he-etal-2025-scaling} or studies bilingual transfer solely in multilingual machine translation~\cite{fernandes2023scaling,DBLP:conf/iclr/Isik00PVK25}. 
Concurrent to our work, ATLAS~\cite{ICLR2026_35c26a81} models cross-lingual transfer via empirically fitted coefficients over a target/transfer/remaining-language decomposition. To the best of our knowledge, ShapleyLaw is the first multilingual scaling law to ground cross-lingual transfer in cooperative game theory: rather than relying on ad hoc fitted coefficients, it derives each language's contribution to target-language performance via SVs, thereby optimizing language-level mixture ratios through a theoretically grounded decomposition of cross-lingual transfer.

\section{Conclusion}


We proposed \textit{ShapleyLaw}, a game-theoretic multilingual scaling law that explicitly incorporates aggregate cross-lingual transfer into multilingual pre-training. Across experiments on ten languages from eight language families, we show that the proposed law accurately predicts per-language pre-training loss across diverse model sizes, data budgets, and language mixtures, while capturing transfer effects that are missed by prior family-level approaches. We demonstrate that ShapleyLaw supports optimization of multilingual sampling ratios, producing better mixtures than heuristic and prior-law baselines under real pre-training. In addition, the predicted language-specific loss was shown to remain informative for downstream multilingual NLU performance, highlighting the practical utility of the framework beyond pre-training loss alone. We also examine the efficiency of approximate Shapley value estimation and show that it preserves strong empirical performance, highlighting the potential scalability of our method to larger multilingual settings. Taken together, these findings suggest that multilingual scaling behavior is better understood through fine-grained, transfer-aware language interactions, and that our method provides a useful foundation for efficient multilingual mixture design.

\newpage

\section*{Limitations}
Due to computational constraints, the largest model scale validated in the main text is $N=1460$M parameters, which already exceeds the largest scale ($N=1.2$B) validated in FamilyLaw~\citep{he-etal-2025-scaling}, the SOTA multilingual scaling law. 
Therefore, we believe that our empirical validation covers a sufficiently broad range of model scales to demonstrate the effectiveness of our method.
To further address potential concerns regarding the extrapolation of ShapleyLaw to larger models, we additionally trained four models at $N=3.2$B parameters. ShapleyLaw continues to predict the test loss of these larger models with high accuracy, providing strong evidence that ShapleyLaw generalizes well beyond the scales used for fitting (see Appendix~\ref{app:large_scale_N}.)

\section*{Ethical Statements}
We anticipate no ethical concerns from this work.
\section*{Use of Generative AI}
AI assistants, such as ChatGPT, were used in this work for grammar checking and language editing.
\section*{Acknowledgments}
This work was financially supported by JST CREST [grant number JPMJCR22M2], JST BOOST [grant number JPMJBS2402], JSPS KAKENHI [grant numbers JP23K17456, JP23K28096, JP25H01117, JP25K21207, JP26K03246], and the United Kingdom Engineering and Physical Sciences Research Council [grant number UKRI5028],
and was computationally supported by mdx and the "Development Acceleration Use" program of ABCI 3.0.


\bibliography{custom}

\newpage 
\clearpage
\newpage
\appendix

\section{More Technical Details}

\subsection{Theoretical Properties of the SV}
\label{app:sv_properties}
The SV uniquely satisfies the following properties.
\begin{itemize}
    \item \textit{Efficiency:} The payoff (i.e., loss reduction) $\payoff_j(\languageset)$ for each task $\task_j$ is fully distributed among all the languages $\languageset$, i.e., $\forall j, \sum_{i=1}^{K}\sv_{i, j}(\mpg) = \payoff_{j}(\languageset)$.
    \item \textit{Linearity:} The SVs of each language for different evaluation tasks are linearly additive.
    Let $\sv_{i, j+j'}$ denotes the Shapley value of language $i$ for task $\task_{j+j'} = \task_j \cup \task_{j'}$. 
    Then, for any language $i$, and for any two tasks $\task_{j}$ and $\task_{j'}$, we have $\sv_{i, j+j'}(\mpg) = \sv_{i, j}(\mpg) + \sv_{i, j'}(\mpg)$.
    That means that the contribution of language $i$ to a combined task $\task_{j+j'}$ should be the sum of its contributions to the individual tasks $\task_{j}$ and $\task_{j'}$.
    \item \textit{Symmetry:} If two languages $i, i'$ are equivalent for improving model performance, then their SVs are identical. 
    Formally, if $\payoff_j(S \cup \{i\}) = \payoff_j(S \cup \{i'\})$ for any language subset $S\subseteq \languageset$, then $\sv_{i,j} = \sv_{i', j}$.
    \item \textit{Null player:} If a language $i$ never improves the model performance when its data is included for pretraining, its contribution should be zero.
    That is, if $\payoff_j(S \cup \{i\}) = \payoff_j(S)$ for any language subset $S\subseteq \languageset$, then $\sv_{i,j} = 0$.
\end{itemize}

\subsection{ShapleyLaw Mixture Optimization}
\label{app:optimal_mixture}

\paragraph{Optimal mixture ratios for language preferences.}
In practice, multilingual pre-training is often guided by an application-specific evaluation suite spanning multiple languages, where different languages may carry different importance.
To model this, suppose we have nonnegative preference weights $\{w_j\}_{j=1}^{K}$ (e.g., proportional to user demand or benchmark priority) and consider the weighted objective
\begin{equation*}
\mathcal{J}(\mathbf{p}) \;=\; \sum_{j=1}^{K} w_j \, \law_{j}(N, D, \ratios)
\label{eq:weighted_objective}
\end{equation*}
where $\law_{j}$ is the test loss for target language $j$.
Introducing ShapleyLaw into Eq.~\ref{eq:weighted_objective} yields
\begin{equation}
\begin{aligned}
\mathcal{J}(\mathbf{p})
\;=\;
\sum_{j=1}^{K} w_j \cdot C_j \cdot \big(\aggregatetransfer_j\big)^{-\gamma_j}
\label{eq:objective_expanded}
\end{aligned}
\end{equation}
where $C_j \triangleq E_j+\frac{A_j}{N^{\alpha_j}}+\frac{B_j}{D^{\beta_j}}$, which makes explicit how mixture influences the weighted loss through $\aggregatetransfer_j$.

\paragraph{Convexity and optimality conditions.}
For $\gamma_j>0$, the function $x\mapsto x^{-\gamma_j}$ is convex over $x>0$; since $\aggregatetransfer_j$ is affine in $\mathbf{p}$, $\mathcal{J}(\mathbf{p})$ is a convex objective over the simplex whenever $\aggregatetransfer_j>0$ for all $j$.
Therefore, the mixture selection problem
\begin{equation*}
\min_{\mathbf{p}\in\Delta^{K-1}} \;\mathcal{J}(\mathbf{p})
\label{eq:mixture_opt}
\end{equation*}
admits a global optimum and can be solved efficiently.
Using a Lagrangian with simplex constraints and Karush-Kuhn-Tucker (KKT) conditions, the gradient of Eq.~\ref{eq:objective_expanded} is
\begin{equation*}
\frac{\partial \mathcal{J}}{\partial p_i}
\;=\;
-\sum_{j=1}^{K} w_j C_j \gamma_j \,
\nsv_{i,j} \, \big(\aggregatetransfer_j\big)^{-(\gamma_j+1)}.
\label{eq:grad}
\end{equation*}
At an interior optimum (all $p_i>0$), these gradients are equal up to a shared Lagrange multiplier, implying an intuitive \emph{marginal utility equalization} principle: optimal mixtures allocate probability mass so that the expected reduction in weighted loss per unit increase of $p_i$ is balanced across languages.

\paragraph{Closed-form solution in a diagonal (self-dominant) approximation.}
A useful and interpretable special case arises when each target language is primarily influenced by its own data, i.e., $\nsv_{i,j}$ is approximately diagonal so that
$\aggregatetransfer_j\approx p_j\,r_j$ with $r_j\triangleq \nsv_{j,j}$.
Then Eq.~\ref{eq:objective_expanded} becomes separable:
\begin{equation*}
\mathcal{J}(\mathbf{p}) \approx \sum_{j=1}^{K} w_j C_j (p_j r_j)^{-\gamma_j}.
\end{equation*}
The KKT conditions yield a closed-form optimal ratio:
\begin{equation}
p_i^{\star}
\;=\;
\frac{\left(w_i C_i \gamma_i \, r_i^{-\gamma_i}\right)^{\frac{1}{\gamma_i+1}}}
{\sum_{m=1}^{K}\left(w_m C_m \gamma_m \, r_m^{-\gamma_m}\right)^{\frac{1}{\gamma_m+1}}}.
\label{eq:closed_form}
\end{equation}

\paragraph{First-order correction under cross-lingual transfer.}
For realistic multilingual settings, cross-lingual transfer introduces non-negligible off-diagonal terms in $\nsv$.
Let the Shapley matrix be decomposed as
\begin{equation*}
\nsv = I + E
\end{equation*}
where $E$ captures cross-lingual interactions with relatively small off-diagonal entries.
Then
\begin{equation*}
\aggregatetransfer = \nsv\mathbf{p} = \mathbf{p} + E\mathbf{p}.
\end{equation*}

Expanding $\aggregatetransfer_i^{-(\gamma_i+1)}$ to first order around the diagonal solution $p^{(0)}$ from Eq.~\ref{eq:closed_form} yields
\begin{equation}
p_i^{\star}
\approx
p_i^{(0)}
\left(
1-
\frac{1}{\gamma_i+1}
\sum_{j,j\neq i} \nsv_{i,j}
\right)
\label{eq:first_order}
\end{equation}
followed by renormalization to satisfy the simplex constraint.
The correction term depends on the column sum $\sum_{j\neq i} \nsv_{i,j}$, which measures how strongly language $i$ contributes to other languages through cross-lingual transfer.
Languages with larger column sums effectively provide training signal to many other languages, and therefore require less direct sampling mass in the optimal mixture.
Conversely, languages with weaker transfer connections receive higher allocation to maintain balanced marginal utility across tasks.

In practice, the available corpus size for each language may be limited.
Let $\bar{D}_i$ denote the maximum usable token budget for language $i$,
and let the total training budget be $D$.
Then the mixture ratio must satisfy the additional box constraint
\begin{equation}
p_i \;\le\; \bar{p}_i
\quad\text{where}\quad
\bar{p}_i \;=\; \frac{\bar{D}_i}{D}.
\label{eq:box_constraint}
\end{equation}
The mixture optimization problem thus becomes
\begin{equation*}
\min_{\mathbf{p}\in\Delta^{K-1}}
\mathcal{J}(\mathbf{p})
\quad
\text{s.t.}\quad
0\le p_i \le \bar{p}_i .
\label{eq:box_opt}
\end{equation*}

Because $\mathcal{J}(\mathbf{p})$ is convex, the constrained problem remains convex and can be characterized via KKT conditions with complementary slackness.
However, a simple and effective approximation can be derived by projecting the unconstrained optimum onto the feasible region.

\paragraph{Clipped closed-form approximation.}
Let $p_i^{(0)}$ denote the unconstrained solution from Eq.~\ref{eq:closed_form}
(or its cross-lingual correction in Eq.~\ref{eq:first_order}).
Define the \emph{active set} of data-limited languages
\[
\mathcal{A}
\;=\;
\{\, i \mid p_i^{(0)} > \bar{p}_i \,\}.
\]
For languages in $\mathcal{A}$, the optimal solution saturates the constraint:
\begin{equation*}
p_i^\star \;=\; \bar{p}_i, \qquad i\in\mathcal{A}.
\end{equation*}

The remaining probability mass is
\[
M \;=\; 1 - \sum_{i\in\mathcal{A}} \bar{p}_i
\]
which is redistributed among the unconstrained languages
$\mathcal{F}=\{1,\dots,K\}\setminus\mathcal{A}$
proportionally to their original optimal ratios:
\begin{equation}
p_i^\star
\;\approx\;
M \cdot
\frac{p_i^{(0)}}{\sum_{m\in\mathcal{F}} p_m^{(0)}},
\qquad i\in\mathcal{F}.
\label{eq:redistribute}
\end{equation}

If redistribution causes additional violations
($p_i^\star>\bar{p}_i$ for some $i\in\mathcal{F}$),
the clipping-and-redistribution step can be repeated.
Since each iteration adds at least one index to $\mathcal{A}$,
the procedure converges in at most $K$ steps.

This approximation corresponds to projecting the unconstrained optimum
onto the truncated simplex defined by Eq.~\ref{eq:box_constraint}.
Intuitively, languages with insufficient data are trained to exhaustion,
while the remaining training budget is allocated among data-rich languages
according to their relative marginal utility.
Notably, the structure of Eq.~\ref{eq:redistribute} preserves the
optimal trade-off encoded by Shapley-aware scaling laws while ensuring
practical feasibility under corpus-size limitations.

\subsection{In-Run Data Shapley for SV Approximation}
\label{app:IRDSsv}

\paragraph{Why In-Run Data Shapley approximation is needed.}
For a multilingual pretraining game (MPG), the Shapley value (SV) of language $i$ on task $j$ is defined as
\begin{align*}
\sv_{i,j}(\mpg) = \textstyle\sum_{S \subseteq \languageset \setminus\{i\}} W(S) \cdot \improvement(S)
\end{align*}
where $W(S) = \frac{|S|!(K-|S|-1)!}{K!}$, and $\improvement(S) = \payoff_{j}(S \cup \{i\}) - \payoff_{j}(S)$.
Exact computation requires evaluating the payoff of all language coalitions, which entails pre-training $2^K-1$ models for $K$ languages.

To avoid repeated pre-training, we replace coalition-level retraining with an In-Run Data Shapley (IRDS) approximation. Instead of asking how the final payoff changes when language $i$ is added to an independently trained coalition, IRDS measures how language $i$ contributes to the validation loss reduction along a specific multilingual pre-training trajectory. Thus, the resulting value directly attributes the cross-lingual transfer observed in the actual training run used for SV
estimation.

\paragraph{Local payoff in IRDS.}
Consider a single multilingual pre-training run with checkpoints $\{w_0,w_1,\dots,w_T\}$. At training step $t$, let $B_t$ be the sampled mini-batch and let $B_{t,i}\subseteq B_t$ denote the subset of examples whose language is $i$. For a language subset $S\subseteq\languageset$, we define a counterfactual one-step update using only examples from languages in $S$:
\begin{align*}
\widetilde{w}_{t+1}(S)
=
w_t
-
\eta_t
\sum_{r\in S}
\sum_{z\in B_{t,r}}
\nabla \ell(w_t,z),
\end{align*}
where $\eta_t$ is the learning rate and $\ell(w,z)$ is the training loss.

For a target validation language $j$, let $\mathcal{V}_j$ be its validation set and let
\begin{align*}
\ell_j(w)
=
\frac{1}{|\mathcal{V}_j|}
\sum_{z^{val}\in \mathcal{V}_j}
\ell(w,z^{val})
\end{align*}
denote the validation loss on task $j$.
We define the local payoff of coalition $S$ at step $t$ as the one-step validation loss reduction:
\begin{align}
u^{(t)}_j(S)
=
\ell_j(w_t)
-
\ell_j(\widetilde{w}_{t+1}(S)).
\label{eq:IRDS_local_payoff}
\end{align}
This local payoff measures how much the current update, if performed using only languages in $S$, improves performance on target language
$j$.

\paragraph{IRDS SV.}
The IRDS SV of language $i$ on task $j$ is obtained by computing the SV of $i$ for each local payoff
$u^{(t)}_j(\cdot)$ and accumulating it across training:
\begin{align}
\hat{\phi}^{IRDS}_{i,j}
=
\sum_{t=0}^{T-1}
\phi_i\!\left(u^{(t)}_j\right).
\label{eq:IRDS_shapley}
\end{align}
By the linearity of SV, this is equivalent to computing the SV of the cumulative payoff
$\sum_{t=0}^{T-1}u^{(t)}_j(\cdot)$.
Therefore, $\hat{\phi}^{IRDS}_{i,j}$ measures the cumulative contribution of language $i$ to the loss reduction on target language $j$ during the same pre-training run.

\paragraph{First-order approximation.}
Directly computing $\phi_i(u^{(t)}_j)$ still requires enumerating language subsets within each step. We therefore approximate the local payoff in Eq.~\ref{eq:IRDS_local_payoff} by a first-order Taylor expansion. Define the validation gradient for target language $j$ as
\begin{align*}
g^{val}_{t,j}
=
\nabla \ell_j(w_t),
\end{align*}
and define the aggregated training gradient of language $i$ at step
$t$ as
\begin{align*}
g_{t,i}
=
\sum_{z\in B_{t,i}}
\nabla \ell(w_t,z).
\end{align*}
Then,
\begin{align*}
u^{(t)}_j(S)
\approx
\eta_t
\sum_{r\in S}
\left\langle g^{val}_{t,j}, g_{t,r}\right\rangle .
\end{align*}
Since the approximated utility is additive over languages, its Shapley value has a closed form:
\begin{align*}
\phi_i\!\left(u^{(t)}_j\right)
\approx
\eta_t
\left\langle g^{val}_{t,j}, g_{t,i}\right\rangle .
\end{align*}
Thus, the first-order IRDS Shapley estimator is
\begin{align}
\hat{\phi}^{IRDS\text{-}1}_{i,j}
=
\sum_{t=0}^{T-1}
\eta_t
\left\langle g^{val}_{t,j}, g_{t,i}\right\rangle .
\label{eq:IRDS_first_order}
\end{align}
Intuitively, language $i$ receives a positive contribution at step $t$ when its training gradient is aligned with the validation gradient of target language $j$, because this update decreases the validation loss on task $j$. Conversely, a negative value indicates that the update from language $i$ increases the validation loss on task $j$.

The first term measures the direct alignment between language $i$ and the target language $j$. IRDS also provides a second-order approximation that captures interactions between language $i$ and the full multilingual batch under the local curvature of the validation loss. In practice, we use the first-order estimator by default because it is simpler, more stable, and incurs minimal additional cost. The second-order estimator can be used when within-batch language interactions are important.

\begin{algorithm}[t]
\small
\caption{In-Run Data Shapley Approximation}
\label{alg:IRDSsv}
\begin{algorithmic}[1]
\Require Language set $\languageset$, uniform multilingual pre-training data,
validation sets $\{\mathcal{V}_j\}_{j=1}^K$, learning rates
$\{\eta_t\}_{t=0}^{T-1}$
\Ensure Approximate IRDS Shapley values
$\{\hat{\phi}^{IRDS}_{i,j}\}_{i,j\in\languageset}$
\For{$i=1$ to $K$}
    \For{$j=1$ to $K$}
        \State $\hat{\phi}^{IRDS}_{i,j}\gets 0$
    \EndFor
\EndFor
\For{$t=0$ to $T-1$}
    \State Sample multilingual mini-batch $B_t$
    \For{$i=1$ to $K$}
        \State $B_{t,i}\gets \{z\in B_t: \mathrm{lang}(z)=i\}$
        \State $g_{t,i}\gets \sum_{z\in B_{t,i}}\nabla \ell(w_t,z)$
    \EndFor
    \For{$j=1$ to $K$}
        \State $g^{val}_{t,j}\gets \nabla \ell_j(w_t)$
        \For{$i=1$ to $K$}
            \State $\Delta_{i,j}^{(t)}
            \gets
            \eta_t \langle g^{val}_{t,j}, g_{t,i}\rangle$
            \Comment{First-order IRDS}
            \State $\hat{\phi}^{IRDS}_{i,j}
            \gets
            \hat{\phi}^{IRDS}_{i,j}+\Delta_{i,j}^{(t)}$
        \EndFor
    \EndFor
    \State Update $w_{t+1}$ using the full multilingual batch $B_t$
\EndFor
\State \textbf{return} $\{\hat{\phi}^{IRDS}_{i,j}\}_{i,j\in\languageset}$
\end{algorithmic}
\end{algorithm}

\paragraph{Normalization.}
After obtaining the IRDS Shapley matrix
$\{\hat{\phi}^{IRDS}_{i,j}\}_{i,j\in\languageset}$, we apply the same normalization as in the main text:
\begin{align*}
\hat{\phi}^{NSV}_{i,j}
=
\exp\left(
\hat{\phi}^{IRDS}_{i,j}
-
\max_{i'}\hat{\phi}^{IRDS}_{i',j}
\right).
\end{align*}
The normalized values are then used as cross-lingual transfer strengths in ShapleyLaw:
\begin{align*}
\Theta_j
=
\sum_{i\in\languageset}
p_i\cdot \hat{\phi}^{NSV}_{i,j}.
\end{align*}

\paragraph{Computational complexity.}
Unlike exact SV computation, IRDS requires only one multilingual pre-training run for SV estimation. During this run, the method accumulates language-level gradient contributions at each update step. The first-order estimator requires gradient dot-products between the validation gradient of each target language and the aggregated training gradient of each source language. These dot-products can be computed efficiently during backpropagation and do not require pre-training additional coalition models. Therefore, the number of pretrained models is reduced from $O(2^K)$ for exact SV to $O(1)$ pre-training runs for IRDS.

\subsection{Clarification on Baseline Selection}
\label{app:clarification_sampling_baselines}

The goal of this paper is to develop a multilingual scaling law for performance prediction and language mixture optimization. Accordingly, appropriate baselines are methods that can optimize language mixtures for large-scale training configurations \emph{from observations made under small-scale training configurations}. Uniform Sampling, Smoothed Sampling, FamilyLaw Optimization, and Empirical Mixtures all satisfy this criterion, and we therefore adopt them as our mixture optimization baselines.
 
In contrast, some recent data mixture optimization methods such as DoReMi~\cite{DBLP:conf/nips/Xie0DDLLLL0Y23} address a different problem: optimizing data mixtures under a fixed data budget.
As a result, they cannot predict optimal mixtures for unseen data scales. 
Since the core contribution of ShapleyLaw is mixture optimization across both data and model scales, a direct comparison with these methods would be unfair and would not evaluate the capability that ShapleyLaw is designed to provide. 
We therefore do not include them as baselines.



\section{Experimental Setup Details}
\label{app:setup}

\begin{table*}[t]
\centering
\small
\setlength{\tabcolsep}{5pt}
\renewcommand{\arraystretch}{1.15}
\begin{tabular}{lcccccc}
\toprule
\textbf{Params} & \textbf{Layers} & \textbf{Hidden} & \textbf{Heads} & \textbf{Context} & \textbf{Embedding} & \textbf{Non-embedding} \\
                &                 & \textbf{size}   &                & \textbf{length}  & \textbf{parameters} & \textbf{parameters} \\
\midrule
150M (50M) & 12 & 512  & 8 & 4{,}096 & 101{,}874{,}688 & 50{,}344{,}448 \\
440M (243M) & 16 & 1{,}024 & 8 & 4{,}096 & 203{,}749{,}376 & 243{,}303{,}424 \\
980M (684M) & 20 & 1{,}536 & 8 & 4{,}096 & 305{,}624{,}064 & 684{,}258{,}816 \\
1.8B (1460M) & 24 & 2{,}048 & 16 & 4{,}096 & 407{,}498{,}752 & 1{,}459{,}718{,}144 \\
3.7B (3171M) & 28 & 3{,}072 & 24 & 4{,}096 & 611{,}248{,}128 & 3{,}171{,}068{,}928 \\
\bottomrule
\end{tabular}
\caption{Model configurations and parameter breakdown.}
\label{tab:model_configs}
\end{table*}

\begin{table}[ht]
  \centering
  \begin{tabular}{llr}
    \hline
    \textbf{Family} & \textbf{Languages} & \textbf{Tokens (B)} \\
    \hline
    Germanic        & English / German   & 373 / 450   \\
    Romance         & French / Spanish   & 340 / 397    \\
    Sino-Tibetan    & Chinese           & 788          \\
    Japonic         & Japanese          & 281       \\
    Korean          & Korean            & 52  \\
    Uralic          & Finnish           & 54  \\
    Slavic          & Croatian          & 29  \\
    Austronesian    & Standard Malay    & 12  \\
    \hline
  \end{tabular}
  \caption{\textbf{Corpora used in our experiments.} The study covers ten languages from eight language families, the available token counts are listed for each language.}
  \label{table:corpus}
\end{table}

\begin{table*}[t]
\centering
\setlength{\tabcolsep}{3.2pt}
\renewcommand{\arraystretch}{1.08}

\begin{tabular}{lcccccccccc}
\toprule
\textbf{Dataset / Language} & {en} & {de} & {fr} & {es} & {zh} & {ja} & {ko} & {fi} & {hr} & {ms} \\
\midrule
\textbf{Common Crawl}  & 61.4\% & 8.3\% & 6.4\% & 6.4\% & 7.2\% & 8.2\% & 1.2\% & 0.5\% & 0.3\% & 0.1\% \\
\textbf{FineWeb-2}    & 80.6\% & 3.3\% & 2.3\% & 2.7\% & 6.7\% & 3.1\% & 0.5\% & 0.4\% & 0.3\% & 0.1\% \\
\bottomrule
\end{tabular}
\caption{
\textbf{Normalized statistics across languages in open-source corpora.}
The statistics are derived from the official Common Crawl statistics and, for FineWeb-2, estimated based on the reported disk sizes~\cite{penedo2025fineweb2pipelinescale}.
Specifically, the disk sizes of the FineWeb-2 subsets for
\{en, de, fr, es, zh, ja, ko, fi, hr, ms\} are
\{33\,TB, 1.51\,TB, 1.11\,TB, 1.32\,TB, 2.42\,TB, 1.50\,TB,
213.43\,GB, 143.03\,GB, 35.91\,GB, 31.94\,GB\}, respectively.
}

\label{tab:breakdown_datasets}
\end{table*}

\paragraph{Corpora.}
We use the English $sample\text{-}350BT$ subset~\footnote{A subset randomly sampled from the whole dataset of around 350B gpt2 tokens (388GB).}
from \emph{FineWeb}~\cite{penedo2024the}, and the German, French, Spanish, Chinese, Japanese, Korean, Finnish, Croatian, and Standard Malay subsets from \emph{FineWeb-2} \textit{v2.0.0}~\cite{penedo2025fineweb2pipelinescale}. The total tokens are summarized in the following Table~\ref{table:corpus}. 

\paragraph{Training Configuration.}
We pre-train multilingual LLMs with Megatron-LM~\cite{megatron-lm} on $8$-node $8\times$H200 clusters. 
We adopt the open-source model architectures and training hyperparameters provided by the \emph{llm-jp} project\footnote{https://huggingface.co/llm-jp}.

Table~\ref{tab:model_configs} summarizes the configurations of \highlighted{five} transformer language models with increasing non-embedding parameter scales (50M, 243M, 684M, 1460M, and 3171M). As the scale grows, the number of layers and hidden dimensions increase, while the number of attention heads and context length remain constant. Larger models allocate substantially more parameters to both embedding and non-embedding components, with non-embedding parameters growing more rapidly, indicating that most additional capacity comes from deeper and wider transformer blocks rather than the embedding layer.


We adopt a decoder-only Transformer with RMSNorm, RoPE, and SwiGLU. Tokenization uses the LLaMA-2–style \verb|llm-jp-tokenizer-100k.ver3.0b1|\footnote{https://github.com/llm-jp/llm-jp-tokenizer}
 tokenizer with a 100k multilingual vocabulary.
Models are trained with learning rate $3\times10^{-4}$, sequence length 4{,}096, and global batch size 512 using Adam ($\beta_1{=}0.9$, $\beta_2{=}0.95$), cosine decay with warmup, and standard stabilization techniques (weight decay, gradient clipping, Z-loss). Training efficiency is improved through $\mathrm{bf16}$ precision, FlashAttention, fused kernels, and Megatron-LM parallelism, \highlighted{thereby} enabling stable large-scale multilingual pre-training.

\paragraph{Mixture Optimization Baselines.}
We consider the following baselines for mixture optimization.
\begin{itemize}
    \item \textit{Uniform Sampling:} All languages are assigned equal proportions. 
    This balanced heuristic serves as the simplest non-parametric baseline.
    \item \textit{Smoothed Sampling~\citep{DBLP:conf/nips/ConneauL19}:} This strategy downweights dominant languages while still reflecting corpus size through a smoothing parameter $\alpha$, i.e., $p_i = \frac{(\corpussize_i/D)^\alpha}{\sum_{i'=1}^{K} (\corpussize_{i'}/D)^\alpha}$.
    We adopt the common setting $\alpha = 0.5$ for all the related experiments. 
    \item \textit{FamilyLaw Optimization~\citep{he-etal-2025-scaling}:} This strategy first optimizes family-level mixture ratios using FamilyLaw and then allocates language-level mixture ratios using the Smoothed Sampling (see Appendix F in~\citep{he-etal-2025-scaling} for details).
    \item \textit{Empirical Mixtures:} \highlighted{As shown in Table~\ref{tab:breakdown_datasets}, empirical mixture ratios are drawn based on statistics of existing open-source corpora, including \emph{Common Crawl}~\footnote{https://commoncrawl.github.io/cc-crawl-statistics/plots/languages.html} and \emph{FineWeb-2}. Note that the available Korean corpus is larger than the Finnish corpus in disk size (213.43,GB vs.\ 143.03,GB) of \emph{FineWeb-2}, whereas the Finnish corpus contains slightly more tokens (54B vs.\ 52B), likely due to our tokenizer.}
    We include this setting to evaluate whether the proposed method improves upon corpus-native sampling strategies.
\end{itemize}


\paragraph{Downstream Tasks.} 
\label{app:downstream_tasks}
We evaluate all seven pre-training languages on publicly available benchmarks covering linguistic competence, natural language understanding, and commonsense reasoning. The tasks are: \emph{HellaSwag} (commonsense completion)~\cite{zellers-etal-2019-hellaswag,lai-etal-2023-okapi}, \emph{XStoryCloze} (cross-lingual story ending prediction)~\cite{DBLP:conf/emnlp/LinMAWCSOGBDPSK22,mostafazadeh-etal-2017-lsdsem}, and \emph{XWinograd} (multilingual coreference resolution)~\cite{DBLP:conf/acl/TikhonovR21}.  
Following standard multilingual evaluation practice, we use a \emph{five-shot} setting and report accuracy.

\section{Additional Experimental Results}
\label{app:experiment_result}

\begin{figure*}[t]
  \centering
  \includegraphics[width=0.48\linewidth]{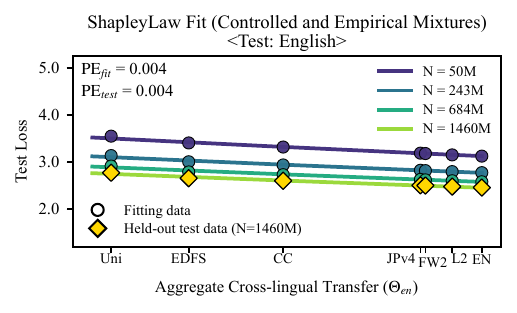}
  \hfill
  \includegraphics[width=0.48\linewidth]{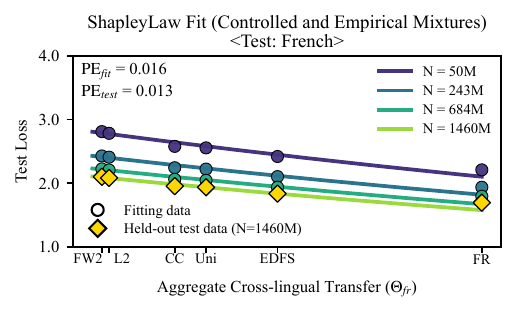}

  \vspace{0.5em}
  
  \includegraphics[width=0.48\linewidth]{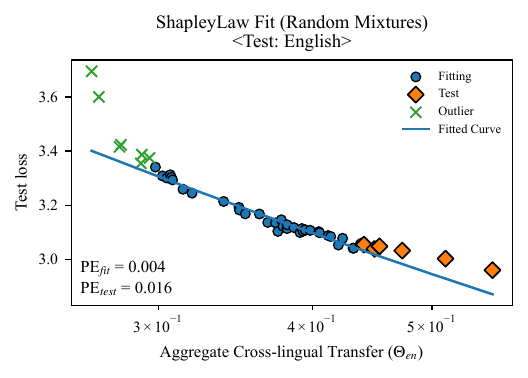}
  \hfill
  \includegraphics[width=0.48\linewidth]{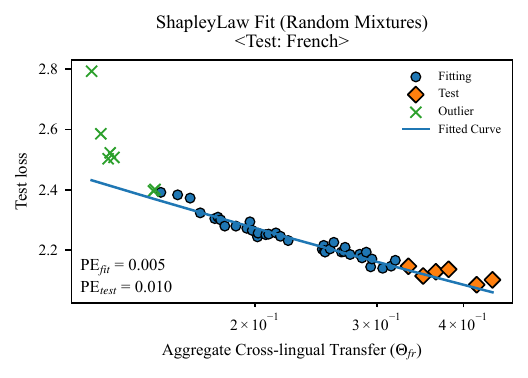}
  
  \caption{\textbf{Additional fitting results for Shapley-based scaling laws on <EN,FR>.}
\textbf{Top row}: Fits on curated multilingual mixtures for \texttt{EN}, and \texttt{FR} across four model scales $\{50\text{M},243\text{M},684\text{M},1460\text{M}\}$; gray circles denote training points and yellow diamonds denote held-out points.
\textbf{Bottom row}: Fits under randomly sampled mixtures under model scale $N$=243M. For both languages, the fitted curves remain smooth and monotonic with respect to $\Theta$, and held-out points closely follow predicted trends, demonstrating strong generalization to unseen mixtures.
}
  \label{fig:extrapolation_en_fr}
\end{figure*}

\begin{figure*}[t]
  \centering
  \includegraphics[width=0.48\linewidth]{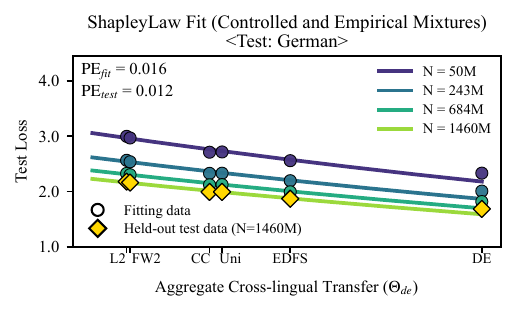}
  \hfill
  \includegraphics[width=0.48\linewidth]{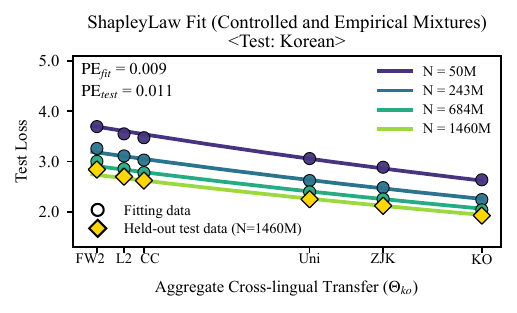}
    
  \vspace{0.5em}
  
  \includegraphics[width=0.48\linewidth]{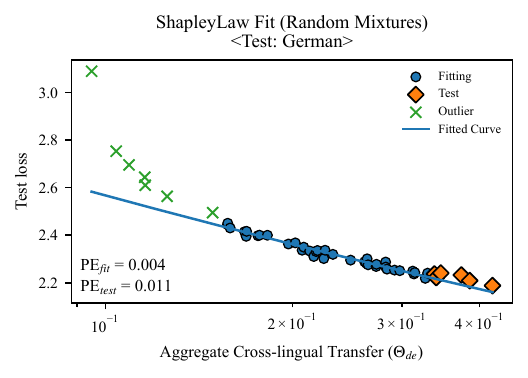}
  \hfill
  \includegraphics[width=0.48\linewidth]{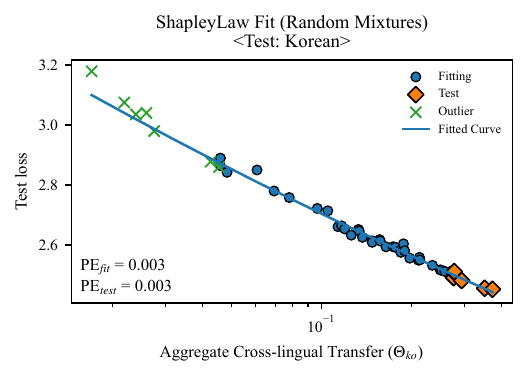}
  
  \caption{\textbf{Additional fitting results for Shapley-based scaling laws on <DE,KO>.} 
  \textbf{Top row}: Fits on curated multilingual mixtures for \texttt{DE} and \texttt{KO} across the same four model scales.
  \textbf{Bottom row}: Fits under randomly sampled mixtures with model scale $N$=243M. For both languages, the fitted curves remain smooth and monotonic with respect to aggregate cross-lingual transfer $\Theta$, and held-out points closely follow predicted trends, demonstrating strong generalization to unseen mixtures.}
  \label{fig:extrapolation_de_ko}
\end{figure*}

\subsection{Additional Experiments on Performance Prediction.}
\label{app:performance_predict}
To further validate the robustness of the proposed Shapley-based scaling law, we report additional fitting results on four target languages: \texttt{EN}, \texttt{FR}, \texttt{DE}, and \texttt{KO}. 
For each target language, we fit the proposed relation using both curated multilingual mixtures and randomly sampled mixtures, and evaluate whether the aggregate cross-lingual transfer $\Theta$ consistently explains the variation in held-out test CE loss.
Representative examples for \texttt{EN}, and \texttt{FR} are shown in Figure~\ref{fig:extrapolation_en_fr}: the top row presents fits across predefined mixture settings and model scales, while the bottom row shows results under random mixture sampling.
In both cases, empirical losses align closely with the predicted curves, and held-out points remain near the fitted trend, indicating that the scaling law generalizes beyond the mixtures used for fitting.

The same pattern is observed for \texttt{DE}, and \texttt{KO} (see Figure~\ref{fig:extrapolation_de_ko}), where losses across different model sizes and sampling strategies can also be well described by the ShapleyLaw.
Across these additional languages, the fitted curves remain smooth and monotonic with respect to $\Theta$, and the resulting high $R^2$ values range from 0.985 to 0.998.
These results provide further evidence that the proposed parameterization is not specific to a single language or mixture family, but instead captures a common regularity in multilingual pre-training.
Overall, the additional results strengthen the conclusion that aggregate cross-lingual transfer serves as an effective mixture variable for modeling per-language scaling behavior across diverse target languages.

\subsection{Additional Experiments on Downstream Tasks}
\label{app:downstream_task}

\begin{table}[t]
\centering
\small
\begin{tabular}{lccccc}
\toprule
Task & $\ell$ & $-\ell$ & $\log \ell$ & $\exp(-\ell)$ & $1/\ell$ \\
\midrule
HellaSwag & 0.994 & 0.994 & 0.995 & \textbf{0.996} & \textbf{0.996} \\
XWinograd & 0.971 & 0.971 & 0.975 & \textbf{0.980} & 0.978 \\
XStorycloze & \textbf{0.947} & \textbf{0.947} & 0.943 & 0.936 & 0.939 \\
\bottomrule
\end{tabular}
\caption{\textbf{Model Selection.} Model comparison for $s = a\cdot g(\ell) + b$ using different transformations of loss. Numbers denote $R^2$. Across tasks except XStorycloze, $g(\ell)=\exp(-\ell)$ consistently achieves the highest explanatory power, suggesting that accuracy scales approximately linearly with token-level probability $p = \exp(-\ell)$.}
\label{tab:model_selection}
\end{table}

\begin{table*}[t]
\centering
\begin{tabular}{lcccccc}
\toprule
Task & Pearson $r$ & $p$-value & Best $g(\ell)$ & $R^2$ & $a$ & $b$ \\
\midrule
HellaSwag & -0.997 & $8.15\times10^{-7}$ & $\exp(-\ell)$ & 0.996 & 4.030 & 0.098 \\
XWinograd & -0.985 & $4.92\times10^{-5}$ & $\exp(-\ell)$ & 0.980 & 4.824 & 0.382 \\
XStorycloze & -0.973 & $2.25\times10^{-4}$ & $\exp(-\ell)$ & 0.980 & 3.034 & 0.433 \\
\bottomrule
\end{tabular}
\caption{\textbf{Fitted Results (English Downstream Tasks).} Pearson correlation between pretraining loss $\ell$ and few-shot accuracy. All tasks exhibit strong negative correlation ($|r| > 0.90$), indicating that lower loss consistently corresponds to higher downstream performance. Correlation between pretraining loss $\ell$ and few-shot accuracy, and best-fit results for $\hat{s} = a g(\ell) + b$. Across all tasks, $\exp(-\ell)$ consistently achieves the highest $R^2$.}
\label{tab:loss_fewshot_fit}
\end{table*}

\begin{figure*}[t]
  \includegraphics[width=0.33\linewidth]{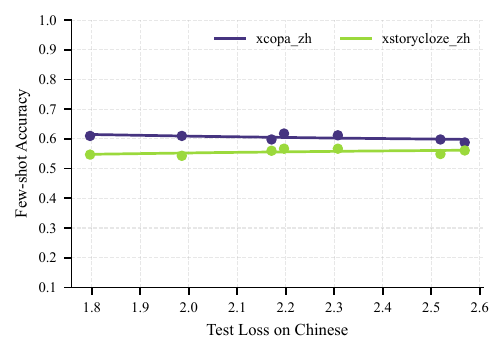}\hfill
  \includegraphics[width=0.33\linewidth]{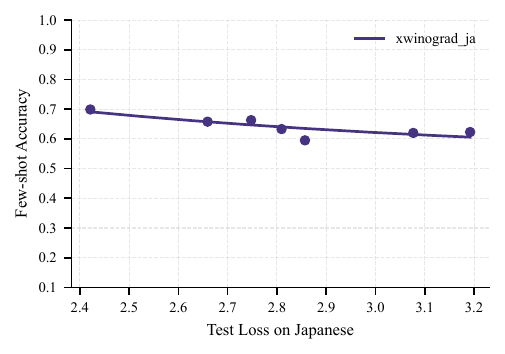}
  \hfill
  \includegraphics[width=0.33\linewidth]{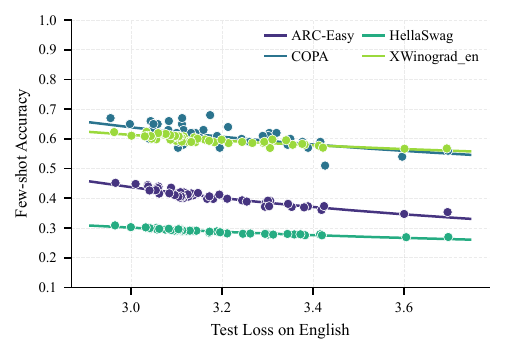}
  \caption {\textbf{Left.} Downstream performance versus test cross-entropy (CE) loss on Chinese tasks (\textit{XCOPA\_zh} and \textit{XStoryCloze\_zh}).
    \textbf{Middle.} Downstream performance versus test CE loss on the Japanese task (\textit{XWinograd}).
    \textbf{Right.} Downstream performance for English versus test CE loss across 50 models ($N=243$M, $D=100$B) trained with randomly sampled data mixtures.} 
  \label{fig:downstream_app}
\end{figure*}

\paragraph{Results of Model Selection.}
To determine an appropriate form for $g(\ell)$, we evaluate five candidate transformations: $\{\ell, -\ell, \log \ell, \exp(-\ell), 1/\ell\}$. For each candidate, we fit a linear model of the form $s = a\cdot g(\ell) + b$ that maps loss values to predicted scores. Table~\ref{tab:model_selection} reports the resulting model comparisons. Among these transformations, $\exp(-\ell)$ always exhibits the strongest explanatory power, achieving the highest $R^{2}$ and consistently providing the best fit to the observed relationship between loss and score. These results indicate that an exponential transformation of loss most effectively captures the empirical mapping between the two quantities, and we therefore adopt $g(\ell)=\exp(-\ell)$ in the subsequent experiments.

\paragraph{Fitting Results on Downstream Tasks.}
We next model the relationship between pre-training loss $\ell$ and downstream performance using the functional form $s = a \cdot \exp(-\ell) + b$. Table~\ref{tab:loss_fewshot_fit} reports the fitting results across six English downstream tasks, comparing pre-training loss with few-shot (5-shot) accuracy. Across all tasks, we observe consistently strong \emph{negative correlations}, with Pearson coefficients ranging from $-0.909$ to $-0.997$ ($p\text{-value} < 0.01$ in all cases), indicating that lower pre-training loss is strongly associated with higher downstream performance.
The fitted models also achieve high explanatory power, with $R^2$ values between 0.843 and 0.996, suggesting that the exponential transformation $\exp(-\ell)$ captures the empirical relationship between loss and accuracy well. These results support the use of $s = a\cdot  \exp(-\ell) + b$ as a simple yet effective mapping for predicting downstream task performance from pre-training loss.

\paragraph{Additional Fitting Results on Other Language Tasks.}
We conduct additional experiments on downstream tasks for Chinese and Japanese to explore the feasibility of ShapleyLaw. In addition, we fit results of downstream performance vs test loss of 50 models ($N=243$M, $D=100$B) with randomly sampled language mixtures. As shown in Figure~\ref{fig:downstream_app}, we report fitting results for Chinese, and Japanese.

\subsection{Additional Experiments on SV Approximation}
\label{app:approx_sv_performance}

\begin{table}[t]
\centering
\small
\begin{tabular}{lcc}
\hline
\textbf{Comparison} & \textbf{Cosine Similarity} & \textbf{Pearson} \\
\hline
Exact SV vs. IRDS SV & 0.9584 & 0.9390 \\
\hline
\end{tabular}
\caption{Similarity comparison between exact SV and IRDS SV.}
\label{tab:sv_similarity_comparison}
\end{table}

\begin{figure*}[t]
    \centering
    \includegraphics[width=0.33\linewidth]{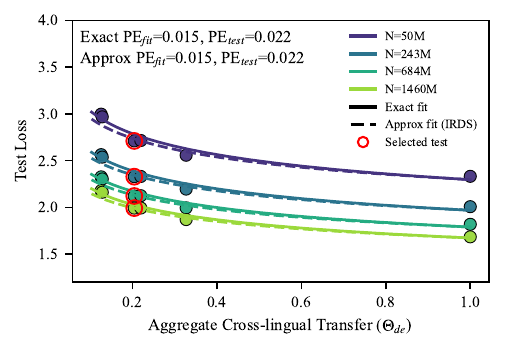}\hfill
    \includegraphics[width=0.33\linewidth]{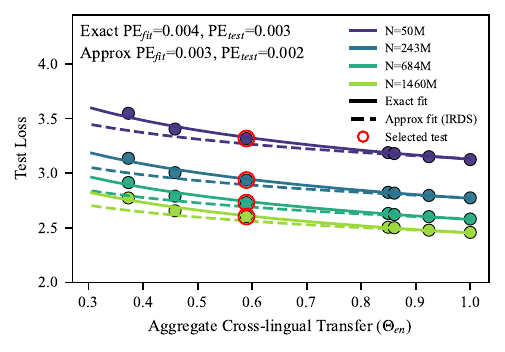}\hfill
    \includegraphics[width=0.33\linewidth]{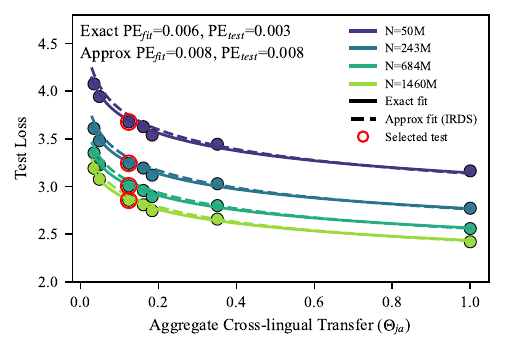}
    \caption{\textbf{Fitting results with exact and IRDS-approximated SVs.}
    To reduce the workload of SV computation, we use the IRDS approximation by pretraining a small-scale model 
    ($N=50$M, $D=50$B) with a uniform mixture ratio over seven languages. The resulting IRDS SVs produce fitting curves that closely match those obtained with exact SVs across target languages.}
    \label{fig:approx_IRDS_fit}
\end{figure*}

\begin{table}[t]
\centering
\footnotesize
\setlength{\tabcolsep}{3.2pt}
\renewcommand{\arraystretch}{1.08}
\begin{tabular}{lccccccc}
\toprule
\textbf{Target} & de & en & fr & es & ja & ko & zh \\
\midrule
\textbf{Exact}  & 0.015 & 0.004 & 0.015 & 0.014 & 0.006 & 0.009 & 0.015 \\
\textbf{IRDS}    & 0.015 & 0.003 & 0.036 & 0.017 & 0.008 & 0.023 & 0.013 \\
\bottomrule
\end{tabular}
\caption{Comparison of corresponding $\mathrm{PE}_{fit}$ values for the fitted results with exact and IRDS SVs.}
\label{tab:approx_IRDS_fit_PE}
\end{table}

\paragraph{Experiment Design.}
Exact SV computation requires evaluating all non-empty language coalitions, resulting in $2^{K}-1$ pretrained models for $K$ languages. This cost grows exponentially with $K$ and becomes prohibitive in large multilingual settings. To assess whether ShapleyLaw remains effective without exact SVs, we evaluate IRDS as a scalable approximation that estimates language contributions from a single model pretrained with a uniform multilingual mixture. We use the same MPG as in the main experiments. Specifically, we pretrain a small-scale model with $N=50$M and $D=50$B using a uniform mixture over seven languages, estimate the IRDS matrix, and construct the aggregate transfer variable $\Theta_j$. We then fit ShapleyLaw under the same settings as the exact-SV variant, isolating the effect of SV approximation on both transfer estimation and scaling-law fitting quality.

\paragraph{Similarity between Exact and IRDS SVs.}
Table~\ref{tab:sv_similarity_comparison} reports the similarity between exact SVs and IRDS-estimated SVs. The two SV matrices are highly aligned, achieving a cosine similarity of $0.9584$ and a Pearson correlation of $0.9390$. This indicates that IRDS preserves the global geometry of the cross-lingual transfer matrix: languages that receive large contributions under exact SV estimation also tend to receive large contributions under IRDS. Therefore, although IRDS does not explicitly enumerate all language coalitions, it provides a close approximation to the relative transfer strengths needed by ShapleyLaw.

\begin{figure*}[t]
  \centering
  \includegraphics[width=0.33\linewidth]{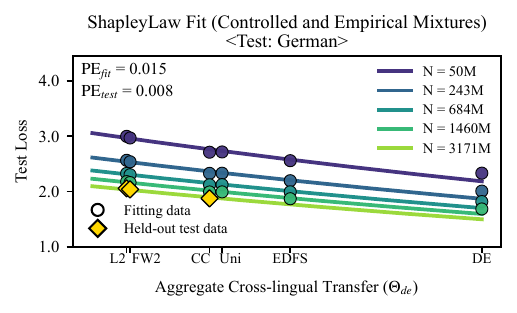}
  \hfill
  \includegraphics[width=0.33\linewidth]{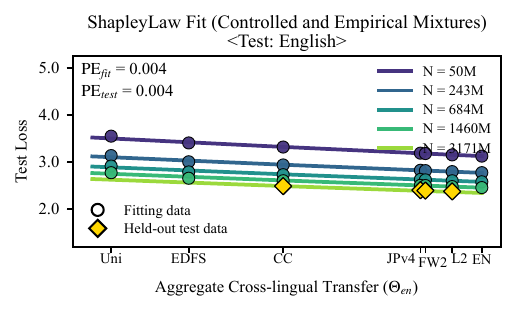}
  \hfill
  \includegraphics[width=0.33\linewidth]{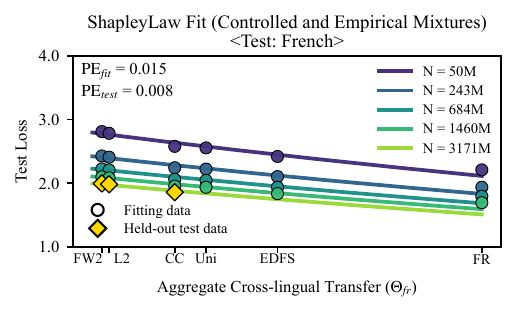}

  \vspace{0.5em}
  
  \includegraphics[width=0.33\linewidth]{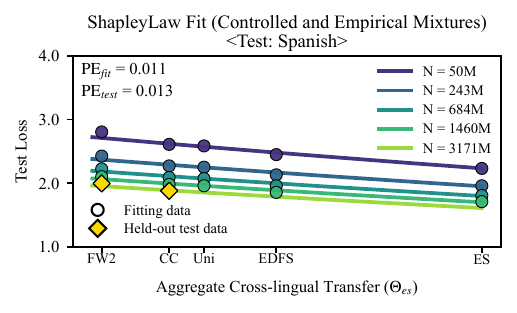}
  \hfill
  \includegraphics[width=0.33\linewidth]{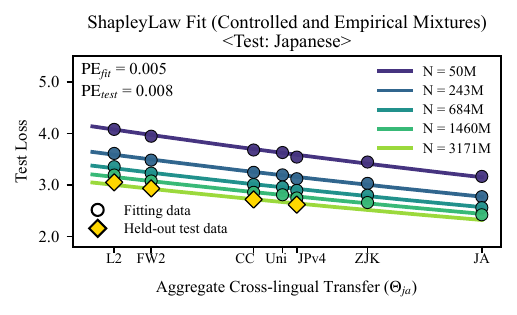}
  \hfill
  \includegraphics[width=0.33\linewidth]{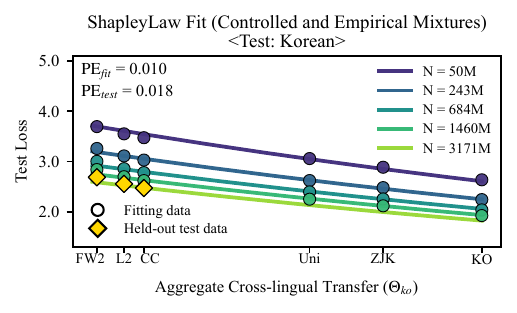}
  
  \caption{\textbf{Additional fitting results for ShapleyLaw at a larger model scale.}
  We extend the evaluation to $N=3171$M and $D=100$B.
  Fits are shown for \texttt{DE}, \texttt{EN}, \texttt{FR}, \texttt{ES}, \texttt{JA}, and \texttt{KO} across 
  model scales $\{50\text{M},243\text{M},684\text{M},1460\text{M},3171\text{M}\}$.
  Gray circles denote fitting points and yellow diamonds denote held-out points.}
  \label{fig:extrapolation_6lang_37}
\end{figure*}

\paragraph{Fitting Results.}
Figure~\ref{fig:approx_IRDS_fit} shows that the IRDS-based fitting curves closely match those obtained with exact SVs across target languages, indicating that ShapleyLaw is robust to approximation noise in the SV matrix. The close alignment of the curves suggests that IRDS preserves this transfer structure sufficiently for loss prediction. Table~\ref{tab:approx_IRDS_fit_PE} further confirms this trend. IRDS achieves fitting errors comparable to exact SVs for most languages, matching or slightly improving the exact-SV results for German, Chinese, and English. The degradation is modest for Japanese and Spanish but more pronounced for French and Korean. Nevertheless, the PEs remain low, and the fitted trends are broadly consistent with the exact-SV results.

\paragraph{Discussion.}
These results show that exact SV computation is not essential for ShapleyLaw's practical effectiveness. In the seven-language setting, exact SV estimation requires $2^{7}-1=127$ pretrained models, whereas IRDS requires only one. Despite this large cost reduction, IRDS achieves SV estimates with high similarity to exact SVs and yields comparable fitting quality for most languages. The remaining gaps mainly reflect approximation noise and do not materially affect the fitted scaling behavior, supporting ShapleyLaw's scalability to larger multilingual settings.

\subsection{Additional Experiments on Larger Model Scale}
\label{app:large_scale_N}

\paragraph{Experiment Design.}
To further evaluate the extrapolation ability of ShapleyLaw beyond the model scales used in the main experiments, we conduct additional experiments at a larger model size. Specifically, we pretrain models with $N=3171$M and $D=100$B under four empirical multilingual mixtures: \textit{CC}, \textit{FW2}, \textit{L2}, and \textit{JPv4}. These mixtures correspond to commonly used multilingual data distributions and provide a realistic evaluation setting for large-scale multilingual pretraining.

We treat all $N=3171$M results as held-out points and fit ShapleyLaw using the smaller-scale observations. This setting directly tests whether the learned scaling relationship can generalize to a larger model scale that is not used for fitting. We evaluate six target languages, including \texttt{DE}, \texttt{EN}, \texttt{FR}, \texttt{ES}, \texttt{JA}, and \texttt{KO}. For each target language, the input variable remains the SV-based aggregate cross-lingual transfer $\Theta_j$, while the scaling law jointly captures the effects of model size, data size, and language mixture.

\paragraph{Results.}
Figure~\ref{fig:extrapolation_6lang_37} shows that ShapleyLaw accurately predicts the held-out $N=3171$M test losses across all six languages. The held-out points generally lie close to the fitted curves, indicating that the learned relationship between aggregate cross-lingual transfer and test loss remains stable when extrapolated to a larger model scale.

ShapleyLaw achieves low held-out prediction errors across languages: $\mathrm{PE}_{test}=0.008$ for German, $0.004$ for English, $0.008$ for French, $0.013$ for Spanish, $0.008$ for Japanese, and $0.018$ for Korean. These values are comparable to the fitting errors obtained on smaller-scale models, suggesting that the proposed law does not merely interpolate among observed configurations but can also provide reliable predictions for larger-scale models.

The results are encouraging because the held-out points combine two sources of distribution shift: they use a larger model size and empirical multilingual mixtures. Despite this harder setting, ShapleyLaw preserves accurate predictions across typologically diverse target languages. This provides evidence that SV-based aggregate transfer is a stable predictor of multilingual test loss and that ShapleyLaw can support practical mixture evaluation at larger model scales without requiring exhaustive large-scale pretraining for every candidate mixture.

\begin{table*}[t]
\centering
\begin{tabular}{lccccccc}
\toprule
Method & de & en & fr & es & ja & ko & zh \\
\midrule 
Signed Normalization
& 0.039 & \textbf{0.004} & 0.037 & 0.023 & 0.029 & 0.042 & 0.037 \\
Min--Max Normalization
& 0.039 & 0.006 & 0.039 & 0.024 & 0.026 & 0.035 & 0.035 \\
Ours
& \textbf{0.016} & \textbf{0.004} & \textbf{0.016} & \textbf{0.011} & \textbf{0.005} & \textbf{0.009} & \textbf{0.013} \\
\bottomrule
\end{tabular}
\caption{\textbf{Comparison of Shapley-value normalization schemes.}
Prediction errors obtained with Signed Normalization, Min--Max Normalization, and the softmax-like normalization adopted in ShapleyLaw across different test languages. Lower values indicate better prediction performance. The proposed normalization achieves the lowest or tied-lowest prediction error for all evaluated languages.}
\label{tab:normalization_comparison}
\end{table*}

\begin{table*}[t]
\centering
\begin{tabular}{lccccccc}
\toprule
& de & en & fr & es & ja & ko & zh \\
\midrule
IRDS Error ($\ell_1$ distance)
& 10.605 & 9.903 & 9.922 & 9.916 & 9.892 & 10.603 & 9.888 \\
$\Delta PE_{\mathrm{fit}}$
& 0.000 & -0.001 & 0.021 & 0.003 & 0.002 & 0.014 & -0.002 \\
\bottomrule
\end{tabular}
\caption{\textbf{IRDS approximation error and the corresponding change in fitting performance across test languages.}
The first row reports the $\ell_1$ distance between the IRDS-approximated and reference Shapley values, while the second row reports the change in fitting error after using the IRDS approximation. Positive values of $\Delta PE_{\mathrm{fit}}$ indicate increased fitting error, whereas negative values indicate a slight improvement.}
\label{tab:irds_fitting_degradation}
\end{table*}

\begin{table*}[t]
\centering
\begin{tabular}{lcccc}
\toprule
Excluded Language
& Germanic/EN
& Germanic/DE
& Romance/FR
& Romance/ES \\
\midrule
$PE_{\mathrm{fit}}$
& 0.006
& 0.004
& 0.006
& 0.009 \\
$PE_{\mathrm{test}}$
& 0.009
& 0.004
& 0.012
& 0.005 \\
\bottomrule
\end{tabular}
\caption{\textbf{Leave-one-language-out evaluation within language families.}
Experiments are conducted with $N=50$M and $D=50$B. For each setting, all fitting mixtures containing the designated language are excluded, and the fitted ShapleyLaw is evaluated on the corresponding held-out within-family mixture.}
\label{tab:leave_one_language_out}
\end{table*}

\subsection{Analysis of the Normalization Scheme in ShapleyLaw}
\label{app:normalization_scheme}

\highlighted{
ShapleyLaw applies a softmax-like transformation to raw Shapley values in order to obtain normalized transfer coefficients that are comparable across different test-language tasks. In particular, the normalization maps the transfer contribution of each language to $(0,1]$ and anchors the self-transfer coefficient of each test language at $1$. This provides a common scale for modeling and comparing cross-lingual transfer patterns across tasks.
\paragraph{Properties of the underlying cooperative game.}
The cooperative game defined in Definition~\ref{def:game} assigns each language an independent and fixed amount of training data. Consequently, languages do not compete for a shared data budget when forming different coalitions. Under this formulation, adding a language that provides no additional benefit produces a zero marginal contribution, rather than a negative contribution induced by resource competition. Empirically, the raw Shapley values obtained in all of our experiments are positive, indicating that negative marginal contributions do not arise in the evaluated setting.
\paragraph{Effect of the normalization.}
The softmax-like transformation does not preserve the classical null-player property after normalization, since every normalized coefficient is strictly positive. This distinction concerns the normalized representation rather than the underlying cooperative game: the null-player property remains well defined at the level of raw marginal contributions and raw Shapley values. In the setting considered by ShapleyLaw, where the observed raw Shapley values are positive, the normalization primarily serves to place transfer effects from different test-language tasks on a common scale while preserving their relative magnitudes.
\paragraph{Comparison with alternative normalization schemes.}
We additionally consider Signed Normalization and Min--Max Normalization, which preserve a zero-valued contribution under normalization. Table~\ref{tab:normalization_comparison} reports the resulting prediction errors across seven test languages. The softmax-like normalization used in ShapleyLaw consistently yields lower prediction error than the alternatives, with the best or tied-best performance for every evaluated language. These results suggest that anchoring self-transfer at $1$ while smoothly rescaling cross-lingual contributions provides a more effective representation for the predictive modeling objective considered in ShapleyLaw.
}

\subsection{Analysis of Language-wise Variation in Fitting Performance}
\label{app:fitting_degradation_analysis}
\highlighted{
We further examine the language-wise variation in fitting performance when Shapley values are approximated using IRDS. Table~\ref{tab:irds_fitting_degradation} reports both the IRDS approximation error and the corresponding change in fitting error for each test language.
}

\highlighted{
The results do not indicate a clear association between the magnitude of the IRDS approximation error and the resulting change in fitting performance. For example, German exhibits the largest approximation error ($10.605$) but essentially no increase in fitting error. Korean has a similarly large approximation error ($10.603$), yet its fitting error increases by $0.014$. Conversely, French shows the largest increase in fitting error ($0.021$), although its approximation error ($9.922$) is close to those of several other languages. These observations suggest that the uneven language-wise changes in fitting performance cannot be directly explained by the IRDS approximation error alone.
}

\highlighted{
A plausible source of this variation is the sensitivity of the fitting procedure to small perturbations in the fitted data points. The absolute fitting errors are already small across the evaluated languages; consequently, minor variations in the approximated Shapley values can induce relatively different changes in $PE_{\mathrm{fit}}$ across individual test languages, even when their overall approximation errors are similar. Importantly, these variations remain small in absolute magnitude and do not exhibit a systematic relationship with the IRDS approximation error.
}

\subsection{Leave-One-Language-Out Generalization within Language Families}
\label{app:leave_one_language_out}

\highlighted{We further evaluate whether ShapleyLaw can generalize to a language that is absent from the fitting set while another language from the same family remains represented. We consider two language families: the Germanic family, consisting of English (EN) and German (DE), and the Romance family, consisting of French (FR) and Spanish (ES). All experiments use $N=50$M and $D=50$B.
}

\highlighted{
For each setting, we designate one language as the held-out language and remove from the fitting set all mixtures containing that language. The remaining language from the same family is retained during fitting. For example, in the \textit{Germanic/EN} setting, all fitting points involving English are excluded, while fitting points involving German remain available. The resulting model is then evaluated on the corresponding within-family DE--EN mixture. The other three settings are constructed analogously.
}

\highlighted{
As shown in Table~\ref{tab:leave_one_language_out}, ShapleyLaw maintains low prediction error across all four leave-one-language-out settings. When EN and DE are held out from the fitting set, respectively, the resulting $PE_{\mathrm{test}}$ values are $0.009$ and $0.004$. Similarly, holding out FR and ES yields $PE_{\mathrm{test}}$ values of $0.012$ and $0.005$, respectively. These values remain close to the corresponding fitting errors of $0.006$, $0.004$, $0.006$, and $0.009$.
}

\highlighted{
These results provide evidence that ShapleyLaw can generalize to a language excluded from the fitting data when another language from the same family remains represented. This within-family generalization suggests that transfer patterns learned from related languages can provide useful predictive signals for languages with limited fitting observations. Broader evaluation across additional language families and more diverse linguistic relationships is required to characterize the generality of this behavior.
}

\end{document}